%% file: main.tex
\documentclass[10pt,journal,compsoc]{IEEEtran}

\ifCLASSOPTIONcompsoc
  \usepackage[nocompress]{cite}
\else
  \usepackage{cite}
\fi
\usepackage[pdftex]{graphicx}
\usepackage{amsmath}
\usepackage{array, boldline, makecell, booktabs}
\usepackage{amssymb}
\usepackage{caption}
\usepackage{acronym}
\usepackage{algorithmic}
\usepackage{breakurl}
\usepackage{subcaption}
\usepackage[edges]{forest}
\usepackage{upgreek}

\usepackage{hyperref}
\hypersetup{
    colorlinks=true,
    linkcolor=blue,
    filecolor=magenta,      
    urlcolor=cyan,
}
\usepackage{enumitem}
\usepackage{multirow}

\usepackage{pifont}
\usepackage{dsfont}

\newcommand{\STAB}[1]{\begin{tabular}{@{}c@{}}#1\end{tabular}}

\newcommand{\ie}{\emph{i.e.} }
\newcommand{\eg}{\emph{e.g.} }

\newcommand{\myparagraph}[1]{\vspace{2pt}\noindent{\bf{#1}}}

\newcommand{\ours}{Co-CGE}
\newcommand{\oursEe}{Co-CGE}
\newcommand{\oursEE}{Co-CGE}

\newcommand{\oursCw}{Co-CGE$^\text{CW}_\text{ff}$}
\newcommand{\oursCwEe}{Co-CGE$^{\text{CW}}$}

\newcommand{\expandednick}{Compositional Cosine Graph Embeddings}

\newcommand{\myvspace}[1]{%
\vspace{#1} %
}

\begin{document}
\title{Learning Graph Embeddings for  \\ Open World Compositional Zero-Shot Learning} 
%
%
%
%

\author{
        Massimiliano~Mancini, 
        Muhammad~Ferjad~Naeem, 
        Yongqin~Xian, 
        and~Zeynep~Akata,~\IEEEmembership{Member,~IEEE}
\IEEEcompsocitemizethanks{\IEEEcompsocthanksitem M. Mancini is with University of
Tübingen.\protect\\
E-mail: massimiliano.mancini@uni-tuebingen.de
\IEEEcompsocthanksitem M.F. Naeem is with ETH Zurich.
\IEEEcompsocthanksitem Y. Xian is with ETH Zurich. The majority of this work was done when Y. Xian was with the Max Planck Institute (MPI) for Informatics.
\IEEEcompsocthanksitem Z. Akata is with University of
Tübingen, MPI for Informatics and MPI for Intelligent Systems.}
\thanks{Manuscript submitted April 29, 2021.}}

\IEEEtitleabstractindextext{
\begin{abstract}
Compositional Zero-Shot learning (CZSL) aims to recognize unseen compositions of state and object visual primitives seen during training. A problem with standard CZSL is the assumption of knowing which unseen compositions will be available at test time. In this work, we overcome this assumption operating on the open world setting, where no limit is imposed on the compositional space at test time, and the search space contains a large number of unseen compositions. To address this problem, we propose a new approach, \expandednick\ (\ours), based on two principles. First, \ours\ models the dependency between states, objects and their compositions through a graph convolutional neural network. The graph propagates information from seen to unseen concepts, improving their representations. Second, since not all unseen compositions are equally feasible, and less feasible ones may damage the learned representations, \ours\ estimates a feasibility score for each unseen composition, using the scores as margins in a cosine similarity-based loss and as weights in the adjacency matrix of the graphs. Experiments show that our approach achieves state-of-the-art performances in standard CZSL while outperforming previous  methods in the open world scenario. 
\end{abstract}

}

\maketitle
\IEEEpeerreviewmaketitle

\IEEEdisplaynontitleabstractindextext

\setcounter{table}{0}
\setcounter{figure}{0}
\ifCLASSOPTIONcompsoc
\IEEEraisesectionheading{\section{Introduction}\label{sec:introduction}}
\else
\section{Introduction}
\label{sec:introduction}
\fi
\input{sections/intro}

\section{Related works}
\input{sections/relateds}

\section{\expandednick}
\label{sec:method}
\input{sections/method}

\section{Experiments}
\input{sections/experiments}

\vspace{2pt}
\section{Conclusions}
\input{sections/conclusion}

\section*{Acknowledgments}
This work has been partially funded by the ERC (853489-DEXIM) and the DFG (2064/1–Project number 390727645).

{\small
\bibliographystyle{IEEEtran}
\bibliography{egbib}
}

\begin{IEEEbiography}
[{\includegraphics[width=1in,height=1.25in,clip,keepaspectratio]{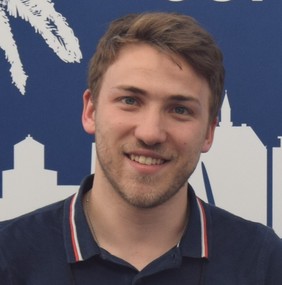}}]{Massimiliano~Mancini}
is a post-doctoral researcher at the Cluster of Excellence in Machine Learning of the University of Tübingen, in the Explainable Machine Learning group, lead by Prof. Zeynep Akata. He completed his PhD in Engineering in Computer Science at the Sapienza University of Rome in 2020. 
During the Ph.D. he has been a member of the ELLIS PhD program, 
the Technologies of Vision lab at Fondazione Bruno Kessler, and the Visual Learning and Multimodal Applications Laboratory of the Italian Institute of Technology. His research interests include transfer learning across domains and learning from low supervision.
\end{IEEEbiography}
\vskip -4em
\begin{IEEEbiography}
[{\includegraphics[width=1in,height=1.25in,clip,keepaspectratio]{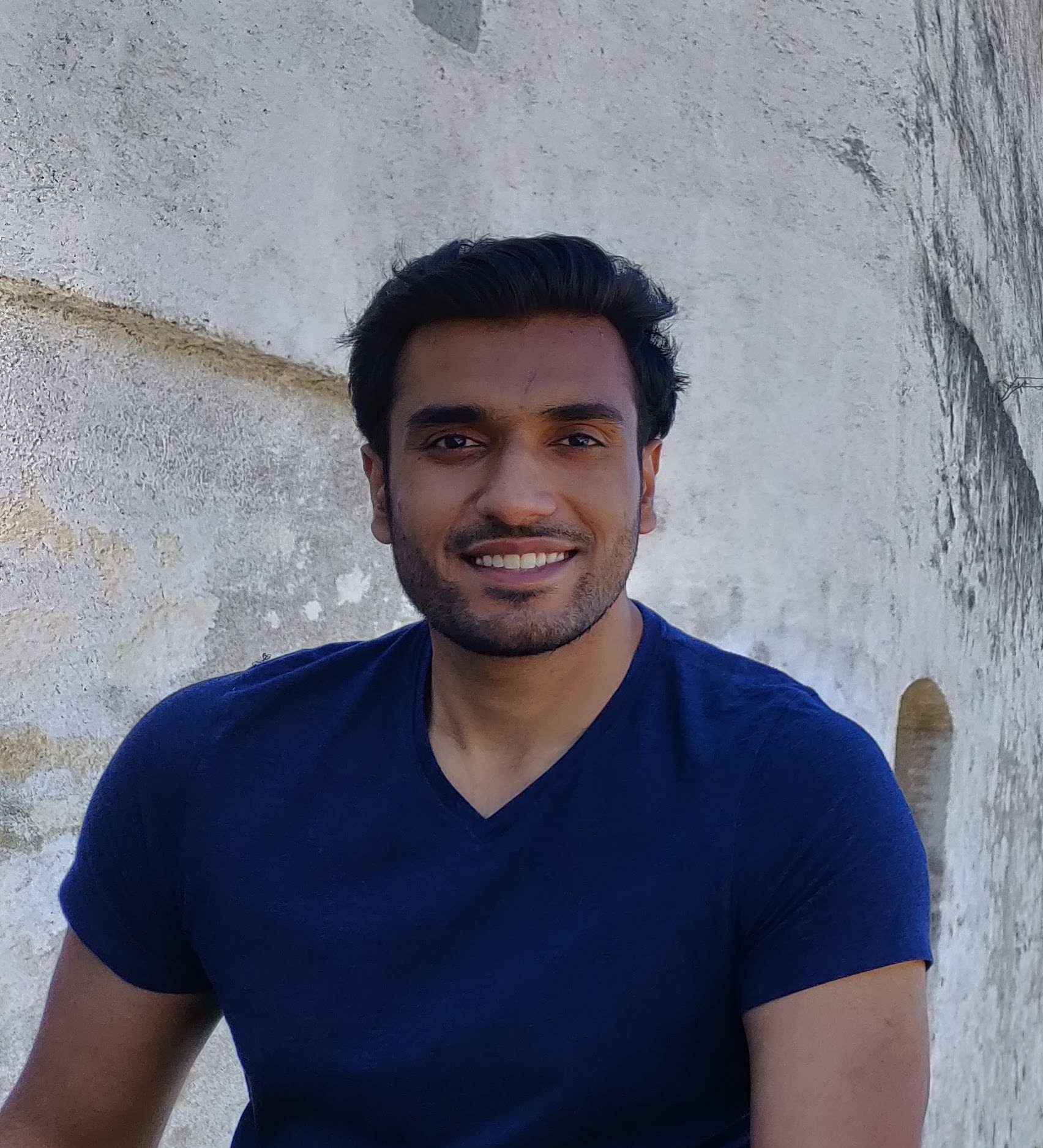}}]{Muhammad~Ferjad~Naeem} is a PhD candidate at the Computer Vision Lab at ETH Zurich supervised by Prof. Luc Van Gool. He completed his Masters at the Technical University of Munich. During his masters, he visited the Explainable Machine Learning group with Prof. Zeynep Akata to work on his master thesis. His Research interests include compositionaility, zero-shot learning and robustness in Machine Learning.
\end{IEEEbiography}
\vskip -4em
\begin{IEEEbiography}[{\includegraphics[width=1in,height=1.25in,clip,keepaspectratio]{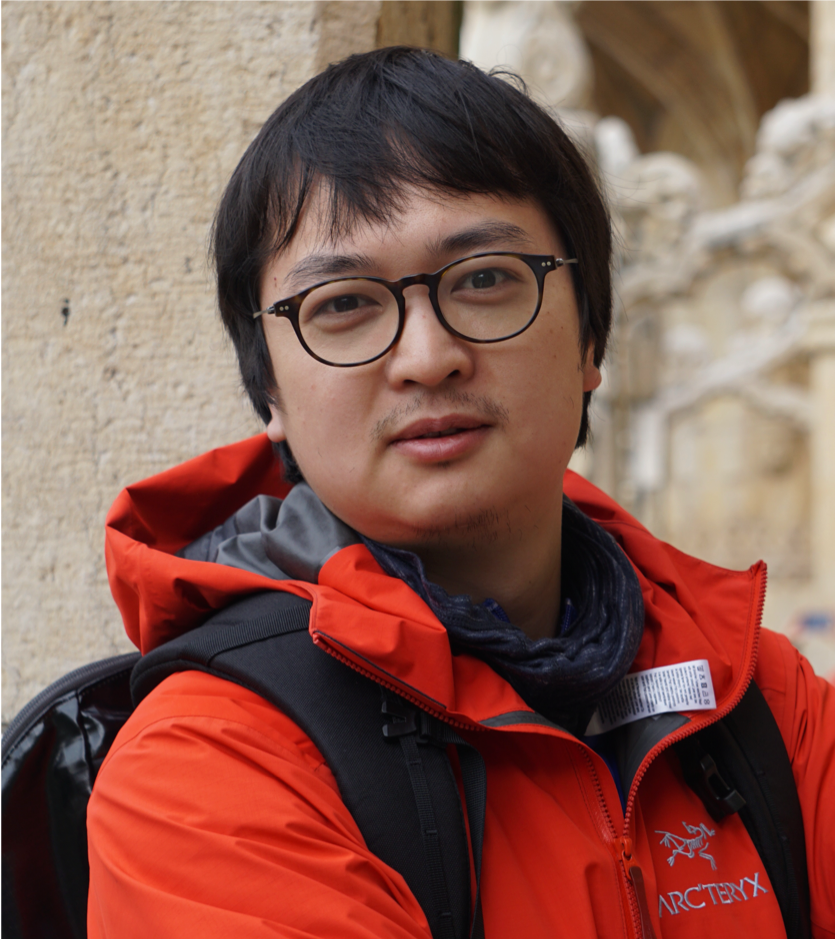}}]{Yongqin Xian}
is a post-doctoral researcher at ETH Zurich, Switzerland. He received a bachelor degree from Beijing Institute of Technology (China) in 2013, a M.Sc. degree with honors from Saarland University (Germany) in 2016 and PhD degree (summa cum laude) from the Max Planck Institute for Informatics (Germany) in 2020. His research interests include zero-shot and few-shot learning for computer vision tasks.
\end{IEEEbiography} 
\vskip -4em
\begin{IEEEbiography}[{\includegraphics[width=1in,height=1.25in,clip,keepaspectratio]{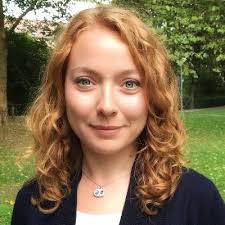}}]{Zeynep Akata} is a professor of Computer Science at the Cluster of Excellence Machine Learning in the University of Tübingen. After her PhD in INRIA Rhone Alpes (F) in 2014, she worked as a post-doctoral researcher at the Max Planck Institute for Informatics (DE) between 2014-2017, at UC Berkeley (USA) between 2016-2017 and as an assistant professor at the University of Amsterdam (NL)  between 2017-2019. She received a Lise-Meitner Award for Excellent Women in Computer Science in 2014 and an ERC Starting Grant in 2019. Her research interests include multimodal learning in low-data regimes and explainable machine learning focusing on vision and language.
\end{IEEEbiography}

\end{document}

%% file: sections/intro.tex
\IEEEPARstart{A} ''black swan'' was ironically used as a metaphor in the 16th century for an unlikely event because the western world had only seen white swans. Yet when the European settlers observed a black swan for the first time in Australia in 1697, they immediately knew what it was. This is because humans posses the ability to compose their knowledge of known entities to generalize to novel concepts. In the literature, this task is known as Compositional Zero-Shot Learning (CZSL)\cite{redwine, aopp, tmn, symnet, cge}, where the goal is to learn how to compose observed objects and their states 
and to generalize to novel state-object compositions. 

CZSL differs from the standard Zero-Shot Learning (ZSL) in multiple aspects. ZSL aims at recognizing unseen categories given an attribute vector describing them. The main objective of ZSL is thus to learn how to map images to their representations in a given, high-dimensional, semantic space. Differently, CZSL assumes access of images containing all primitive concepts (\ie objects and states) but not all their possible compositions. The goal of CZSL is thus to model how states modify objects, extrapolating this knowledge from seen to unseen compositions of the same set of objects and states. 

{A fundamental limitation of both standard ZSL and CZSL is the assumption of operating in a "closed-world", where we know a priori the test-time unseen semantic concepts 
and we can restrict the output space of our model accordingly. As an example, in MIT states there are 115 states and 245 objects, but only 1662 compositions out of 28175 possible combination of these states and objects are considered in the output space at test time. This assumption is unrealistic in practical, unconstrained settings, where arbitrary object categories/compositions may appear. Previous works showed that removing this assumption in ZSL, \ie by considering the whole vocabulary as unseen class set, 
is impractical and leads to poor results \cite{fu2019vocabulary,zslxian18benchmark}. In this paper, we investigate this problem in CZSL, showing that we can consider all possible compositions in the output space, with a relatively small drop in performance despite having a comparable or larger search space (\eg 280k for C-GQA) than the ZSL counterpart (\eg 310k \cite{fu2019vocabulary} or 21k \cite{zslxian18benchmark} for ImageNet). We name this more realistic and challenging setting as Open World CZSL (OW-CZSL).}

{
As in CZSL, in OW-CZSL during training we have images depicting a set of objects in a set of states and, at test time, we receive both seen and unseen compositions of the same objects and states. However, differently from CZSL, in OW-CZSL we assume that we do not know the subset of unseen compositions contained in the test images, thus the model has an output space containing 
\textit{all} possible compositions of the known objects and states (\eg almost 28k compositions in MIT states vs 1662 of standard CZSL).}
Addressing OW-CZSL requires building a representation space where seen and unseen compositions can be recognized despite the huge number of test compositions.  

In this work, we tackle the OW-CZSL task with \expandednick\ (\ours), a graph-based approach for OW-CZSL. \ours\ is based on two inductive biases. Our first inductive bias is a rich dependency structure of different states, objects and their compositions, \eg learning the composition \texttt{old dog} is not only dependent on the state \texttt{old} and object \texttt{dog}, but may also require being aware of other compositions like \texttt{cute dog}, \texttt{old car}, etc. We argue that such dependency structure provides a strong regularization, while allowing the network to better 
generalize to novel compositions and model it through a compositional graph, connecting state, objects and their compositions. 
Differently from previous works \cite{symnet,aopp,tmn,redwine} that treat each state-object composition independently, our graph formulation allows the model to learn compositional embeddings that are globally consistent.

Our second inductive bias is the presence of \textit{distractors}, \ie less feasible compositions (\eg \textit{ripe dog}) that a model needs to either eliminate or isolate in the search space. For this purpose, we use similarities among primitives embeddings to assign a feasibility score to each unseen composition. We then use these scores as margins in a cosine-based cross-entropy loss, showing how the feasibility scores enforce a shared embedding space where unfeasible distractors are discarded, while visual and compositional domains are aligned. Since the distractors may pollute the learned representations of other unseen compositions in \ours, we inject the feasibility scores also within the graph. In particular, we instantiate a weighted adjacency matrix, where the weights depend on the feasibility of each composition. 
Experiments show that \ours\ is either superior or competitive with the state of the art in CZSL while being much more effective on the challenging OW-CZSL task.   

Our contributions are as follows: (1) We introduce \ours, a graph formulation for the new OW-CZSL problem with an integrated feasibility estimation mechanism used to weight the graph connections;  (2) We exploit the dependency between visual primitives and their compositional classes and propose a multimodal compatibility learning framework that embeds related states, objects and their compositions into a shared embedding space learned through cosine logits and feasibility-based margins; (3) We improve the state-of-the-art on MIT states \cite{mitstates}, UT Zappos \cite{utzappos1} and the recently proposed C-GQA \cite{cge} benchmarks on both CZSL and OW-CZSL.

This paper extends our previous works \cite{cge} and \cite{compcos} published in CVPR 2021 in many aspects. First, while being effective on standard CZSL, the CGE model of \cite{cge} performs poorly in the OW-CZSL, due to the noisy connections arising from the huge search space. We thus take the idea of estimating the feasibility of each composition from \cite{compcos} and we inject the feasibility scores both at the loss level and within the graph connections.
Our model is based on a graph convolutional neural network (GCN) \cite{gcn}. This means that the embeddings as well as the feasibility scores are influenced by all other composition in the search space, rather than considered in isolation as it was the case in \cite{compcos}. We extend our OW-CZSL benchmark proposed in \cite{compcos} to the new C-GQA dataset proposed in \cite{cge} {with 413 states, 674 objects thus a total OW-CZSL search space of almost 280k compositions.} Finally, 
we significantly improve the state of the art on the challenging OW-CZSL setting. 

%% file: sections/relateds.tex
Compositionality can loosely be defined as the ability to decompose an observation into its primitives. These primitives can then be used for complex reasoning. One of the earliest attempts in computer vision in this direction can be traced to Hoffman~\cite{hoffman1984parts} and  Biederman~\cite{biederman1987recognition} who theorized that visual systems can mimic compositionality by decomposing objects to their parts.
Compositionality at a fundamental level is already included in modern vision systems. 
Convolutional Neural Networks (CNN) have been shown to exploit compositionality by learning a hierarchy of features\cite{zeiler2014visualizing, lecun1989backpropagation}. Transfer learning\cite{caruana1997multitask, choi2013adding, deng2014large, patricia2014learning}  and few-shot learning\cite{hariharan2017low, ravi2016optimization, mensink2012metric} exploit the compositionality of pretrained features to generalize to data constraint environments. Visual scene understanding\cite{johnson2015image, dai2017detecting, jae2018tensorize, lu2016visual} aims to understand the compositionality of concepts in a scene.
Nevertheless, these approaches still requires collecting data for new compositional classes.

\myparagraph{ZSL and CZSL.} Zero-Shot Learning (ZSL) aims to recognize novel classes not observed during training~\cite{LNH13} using side information describing novel classes \eg attributes~\cite{LNH13}, text descriptions~\cite{RALS16} or word embeddings~\cite{SGMN13}. Some notable approaches include learning a compatibility function between image and class embeddings~\cite{akata2013label, zhang2016learning} and learning to generate image features for novel classes~\cite{xian2018feature, Zhu_2018_CVPR}. 

Compositional Zero-Shot Learning (CZSL) 
aims to learn the compositionality of objects and their states from the training set and generalizing to unseen combinations of these primitives.
Approaches in this direction can be divided into two groups. The first group is directly inspired by \cite{hoffman1984parts, biederman1987recognition}. Some notable methods include learning a transformation upon individual classifiers of states and objects~\cite{redwine}, modeling each state as a linear transformation of objects~\cite{aopp}, learning a hierarchical decomposition and composition of visual primitives\cite{yang2020learning} and modeling objects to be symmetric under attribute transformations\cite{symnet}.
The second group argues that compositionality requires learning a joint compatibility function with respect to the image, the state and the object\cite{causal, tmn, wang2019task}. This is achieved by learning a modular networks conditioned on each composition \cite{tmn,wang2019task} that can be ``rewired" for a new compositions. Finally a recent work from Atzmon et al. \cite{causal} argue that achieving generalization in CZSL requires learning the
visual transformation through a causal graph where the latent representation of primitives are independent of each other.

\myparagraph{GCN.} Graph Convolutional Networks (GCN)~\cite{gcn, gcnzs, gcnzsrethinking} are a special type of neural networks that exploit the dependency structure of data~(nodes) defined in a graph. 
Current methods~\cite{gcn} are limited by the network depth due to over smoothing at deeper layers of the network. The extreme case of this can cause all nodes to converge to the same value~\cite{li2018deeper}. Several works have tried to remedy this by dense skip connections~\cite{xu2018representation, li2019deepgcns}, randomly dropping edges~\cite{rong2019dropedge} and applying a linear combination of neighbor features~\cite{wu2019simplifying, klicpera2019diffusion, klicpera2018predict}. {Recent works in this direction combined residual connections with identity mapping \cite{gcnii} or used .} GCNs have shown to be promising for zero-shot learning. Wang et al.~\cite{gcnzs} propose to directly regress the classifier weights of novel classes with a GCN operated on an external knowledge graph~(WordNet~\cite{wordnet}). Kampffmeyer et al.\cite{gcnzsrethinking} improve this formulation by introducing a dense graph to learn a shallow GCN as a remedy for the Laplacian smoothing problem~\cite{li2018deeper}. 

Our method lies at the intersection of several discussed approaches. We learn a joint compatibility function similar to \cite{causal, tmn, wang2019task} and utilize a GCN similar to \cite{gcnzs, gcnzsrethinking}. However,  we exploit the dependency structure between states, objects and compositions which has been overlooked by previous CZSL approaches~\cite{causal, tmn, wang2019task}. 
Instead of using a predefined knowledge graph like WordNet~\cite{wordnet}, 
we propose a novel way to build a compositional graph and learn classifiers for all classes. 
In contrast to \cite{causal} we explicitly promote the dependency between all primitives and their compositions in our graph. This allows us to learn embeddings that are consistent with the whole graph. Furthermore, our approach estimates the feasibility of each composition, exploiting this information to re-weight the graph connections and to model the presence of distractors within the training objective. Finally, unlike all existing methods~\cite{redwine, aopp, causal, tmn, wang2019task, yang2020learning}, instead of using a fixed image feature extractor our model is trained end-to-end.

\myparagraph{Open World Recognition.}
In our open world setting, all the combinations of states and objects can form a valid compositional class. 
{This is different from an alternate definition of \textit{Open World Recognition} (OWR) \cite{bendale2015towardsowr,mancini2019knowledge} where the goal is to dynamically update a model trained on a subset of classes to detect unknon semantic concepts and incrementally them as new data arrives. Differently from \cite{bendale2015towardsowr} we assume a static set of objects and states, with our open world being the set of their all possible compositions.} 

Our definition is related to the \textit{open set} zero-shot learning (ZSL) \cite{zslxian18benchmark} scenario in \cite{fu2016semi,fu2019vocabulary}, proposing that expands the output space to a large vocabulary of semantic concepts. 
Both our work and 
\cite{fu2019vocabulary}  
consider the lack of constraints in the output space for unseen concepts as a requirement for practical (compositional) ZSL methods.
However, 
since we focus on the CZSL task, we have access to images of all primitives during training but not all their possible compositions. {Note that this differs from \cite{fu2016semi,fu2019vocabulary}, since i) the set of objects and states are fixed and do not vary between training and test time, and ii) we have access to images of all primitives during training but not all their possible compositions, as in standard CZSL.} {From the latter, we can use the knowledge derived from the visual world to model the feasibility of compositions and modifying the representations in the shared visual-compositional embedding space. In this work, we explicitly model the feasibility of each unseen composition, incorporating this knowledge into our model into training.} 

%% file: sections/method.tex
Let $\mathcal{S}$ 
be the set of possible states, with $\mathcal{O}$ being the set of possible objects, and with $\mathcal{C}=\mathcal{S}\times\mathcal{O}$ being the set of all their possible compositions. 
$\mathcal{T}=\{(x_i,c_i)\}_{i=1}^N$ is a training set where $x_i\in\mathcal{X}$ is a sample in the input (image) space $\mathcal{X}$ and $c_i\in\mathcal{C}^s$ is a composition in the subset $\mathcal{C}^s\subset\mathcal{C}$. {In this formulation, $\mathcal{C}^s$ denotes the set of seen compositions.} 
$\mathcal{T}$ is used to train a model $f:\mathcal{X}\rightarrow \mathcal{C}^t$ predicting combinations in a space $\mathcal{C}^t\subseteq\mathcal{C}$ where $\mathcal{C}^t$ may include compositions not present in $\mathcal{C}^s$~(\ie $\exists c \in \mathcal{C}^t \land c\notin \mathcal{C}^s$).

The CZSL task entails different challenges depending on the extent of the target set $\mathcal{C}^t$. If $\mathcal{C}^t$ is a subset of $\mathcal{C}$ and $\mathcal{C}^t\cap \mathcal{C}^s \equiv \emptyset$, the task definition is of \cite{redwine}, where the model needs to predict only unseen compositions at test time. In case $\mathcal{C}^s\subset\mathcal{C}^t$ we are in the generalized CZSL scenario, and the output space of the model contains both seen and unseen compositions. Similar to the generalized ZSL \cite{zslxian18benchmark}, GCZSL scenario is more challenging due to the natural prediction bias of the model in $\mathcal{C}^s$, seen during training. Most recent works on CZSL consider the GCZSL scenario \cite{tmn,symnet}, and the set of unseen compositions in $\mathcal{C}^t$ is known a priori. 

\begin{figure*}[t]
 \includegraphics[width=\linewidth]{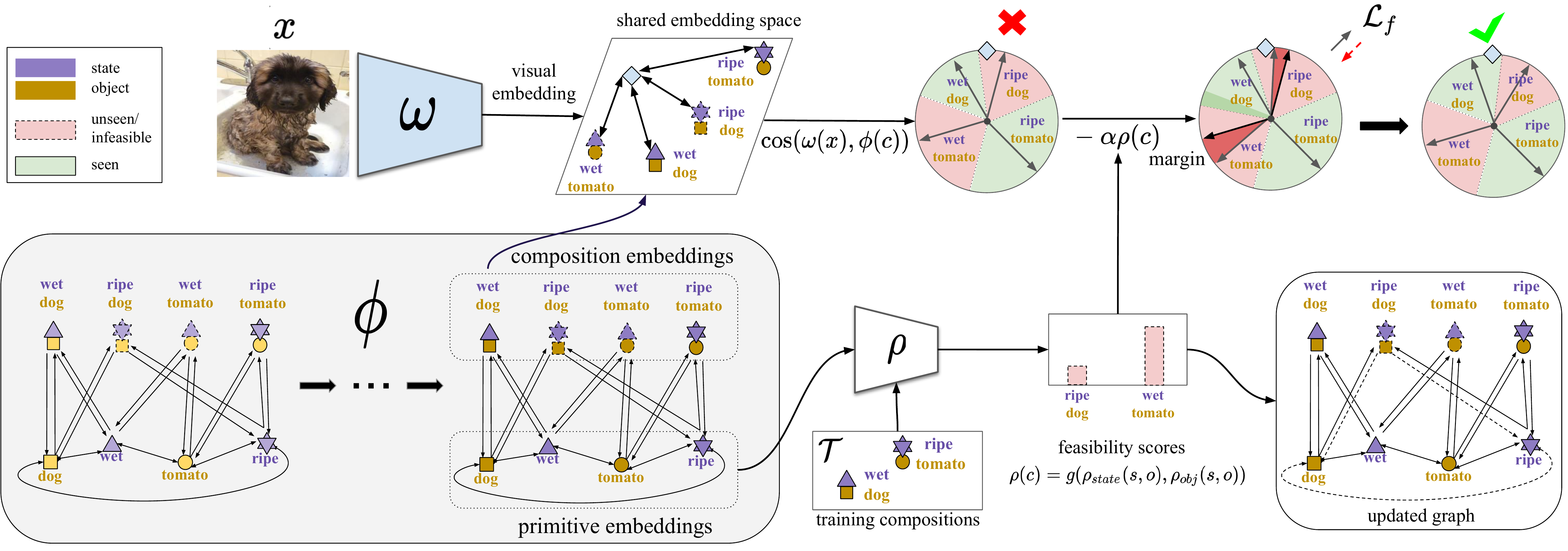}
\vspace{-15pt}
\caption{\textbf{\expandednick\ (\ours).}  Our approach embeds an image (top) and state-object compositions (bottom) into a shared semantic space. The state-object compositions are encoded through a graph (bottom, grey block). 
For OW-CZSL, we estimate a feasibility score for each of the unseen compositions, using the relation between states, objects, and the training compositions. The feasibility scores are injected as margins in the objective function (top right, purple slices) or to update the graph connections (bottom right, dashed lines). 
}
\vspace{-10pt}
\label{fig:method}
\end{figure*}

In our work, the output space is the whole set of possible compositions $\mathcal{C}^t\equiv\mathcal{C}$, \ie \textit{Open World Compositional Zero-shot Learning} (OW-CZSL). Note that this task presents the same challenges of the GCZSL setting while being far more difficult since i) $|\mathcal{C}^t|\gg|\mathcal{C}^s|$, thus it is hard to generalize from a small set of seen to a very large set of unseen compositions; and ii) there are a large number of \textit{distracting} compositions in $\mathcal{C}^t$, \ie compositions predicted by the model but not present in the actual test set that can be close to other unseen compositions, hampering their discriminability. We highlight that, despite being similar to Open Set Zero-shot Learning \cite{fu2019vocabulary}, we do not only consider objects but also states. Therefore, this knowledge can be exploited to identify unfeasible distracting compositions (\eg \textit{rusty pie}) and isolate them.  In the following we describe how we tackle this problem by means of compositional graph embeddings.

\subsection{Compositional Graph Embedding for CZSL }
\label{sec:compcos-closed}

In this section, we focus on the closed world setting, where $\mathcal{C}^s\subset\mathcal{C}^t\subset\mathcal{C}$. Since in this scenario $|\mathcal{C}^t|\ll|\mathcal{C}|$ and the number of unseen compositions is usually lower than the number of seen ones, 
this problem presents several challenges. 
In particular, while learning a mapping from the visual to the compositional space, the model needs to avoid being overly biased toward seen class predictions. 

As states and objects are not independent \eg the appearance of the state \texttt{sliced} varies significantly with the object (\eg \texttt{apple} or \texttt{bread}) and learning state and object classifiers separately is prone to overfit to labels observed during training. Therefore, we model the states and objects jointly via  $f: \mathcal{X} \times \mathcal{C}\rightarrow \mathbb{R}$ that
learns the compatibility between an image and a state-object composition. Given a specific input image $x$, we predict its label $c^*=(s^*, o^*)$ as the state-object composition that yields the highest compatibility score:

\begin{equation}
\label{eq:prediction}
c^* = \arg\max_{c\in\mathcal{C}^t} f(x,c) =  \arg\max_{c\in\mathcal{C}^t}h(\omega(x),\phi(c))
\end{equation}

where $\omega:\mathcal{X}\rightarrow\mathcal{Z}$ is the mapping from the image space to the $d$-dimensional shared embedding space $\mathcal{Z}\in \mathbb{R}^d$, $\phi:\mathcal{C}\rightarrow\mathcal{Z}$ embeds a composition to the same shared space and $h:\mathcal{Z}\times\mathcal{Z}\rightarrow\mathbb{R}$ is a compatibility scoring function. 

We implement $\omega$ as a deep neural network, 
$\phi$ as a graph convolutional neural network (GCN) \cite{gcn} and $h$ as cosine similarity. This way 
we exploit deep image representations and propagate information from seen to unseen concepts through a graph and while at the same time avoid bias on seen classes through cosine similarity scores. We name our model Compositional Cosine Graph Embeddings (\ours). 

\myparagraph{Compositional Graph Embeddings (CGE).} We encode the dependency structure of states, objects and their compositions (both seen and unseen) through a compositional graph. 
We map compositions into the shared embedding space by modeling  
$\phi$ as a GCN  
with $K$ nodes, $L$ layers and the output of the $l_\text{th}$ layer: 
\begin{equation}
    V_{l+1} = \sigma (\hat{A} V_{l} W_l)
    \label{eq:gcn}
\end{equation}
where $\sigma$ is a non-linear activation function (\ie ReLU), $V_l\in \mathbb{R}^{K\times D}$ is the matrix of the $D$-dimensional node representations at layer $l$, 
$W_l\in \mathbb{R}^{D\times D'}$ is the trainable weight matrix at layer $l$ and $\hat{A}\in \mathbb{R}^{K\times K}$ is the column normalized adjacency matrix $A$.
In the CZSL task, CGE defines the set of starting nodes as $V_0 \in \mathbb{R}^{K\times m}$, with $\mathcal{K}=\mathcal{C}^t\cup \mathcal{S}\cup \mathcal{O}$ and $K=|\mathcal{K}|$. Given $k_i$, \ie the i$_\text{th}$ element of $\mathcal{K}$, the representation of its node in $\mathcal{V}_0$ is:
\begin{equation}
    \label{eq:cge-nodes}
    k_i = 
    \begin{cases}
    \varphi(k_i) & \text{if}\,\,\, k_i \in \mathcal{S}\cup\mathcal{O}\\
    (\varphi(s)+\varphi(o))/2 & \text{if}\,\,\, k_i \in \mathcal{C}^t \wedge k_i = (s,o)\\
    \end{cases}
\end{equation}
where $\varphi:\mathcal{S}\cup \mathcal{O}\rightarrow \mathbb{R}^m$ maps the primitives, \ie objects and states, into their corresponding $m$-dimensional embedding vectors.  
The input embeddings of a composition is initialized as the average of its primitive embeddings. All the node representation in $\mathcal{V}_0$ are fixed and initialized with word embeddings, \eg \cite{word2vec}.  A crucial element of the graph is the adjacency matrix $A$. CGE connects all states/objects to objects/states that form at least one composition in the dataset, all composition to their corresponding primitives, and vice-versa. Formally, for two elements $k_i,k_j\in \mathcal{K}$, the value of the adjacency matrix at row $i$, column $j$ is: 
\begin{equation}
    \label{eq:cge-adj}
    A_i^j =
        \begin{cases}
    \textcolor{black}{1} & \textcolor{black}{\text{if} \,\,\,(k_i,k_j)\in\mathcal{C}^t,}\\
    \textcolor{black}{1} & \textcolor{black}{\text{if} \,\,\, (k_j,k_i)\in\mathcal{C}^t,}\\
    1 & \text{if} \,\,\,k_j \in k_i\; \wedge \; k_i \;\in\mathcal{C}^t,\\
    1 & \text{if} \,\,\,k_i \in k_j \; \wedge \; k_j \in\mathcal{C}^t,\\
    1 & \text{if} \,\,\, i=j,\\
    0 &\text{otherwise}\\
    \end{cases}
\end{equation}
where $k_j\in k_i$ is true if $k_i=(k_j,o) \vee k_i=(s, k_j)$. The first case in Eq.~\eqref{eq:cge-adj} denotes the connection between states and objects belonging to the same composition, while the second and the third rows denote the connections between compositions and their base primitives. We highlight that this formulation allows the model 
to propagate the information through the graph, obtaining better node embeddings for both the seen and unseen compositional labels. For example, the GCN allows an unseen composition \eg \textit{old dog} to aggregate information from its seen neighbor nodes \eg  \textit{old}, \textit{dog}, \textit{cute dog}, and \textit{old car}.  {Note that the operation performed to obtain the embeddings is the same of a standard GCN. However, we build the graph nodes and edges in a compositional manner, exploiting our inductive bias (\ie importance of connecting states, objects and their compositions) that is specific for CZSL.}

\myparagraph{Objective function.} The final element of our model is the compatibility score $h$. 
We implement $h$ as the cosine similarity between the visual and compositional embeddings: 
\begin{equation}
\label{eq:cosine}
h(y,z) = \cos(y,z) = \frac{y^\intercal z}{||y||\,||z||}
\end{equation}
to produce bounded scores and it is beneficial to avoid prediction to be influenced by the higher magnitude of scores for seen training classes \cite{hou2019learning} while generalizing better to new ones \cite{gidaris2018dynamic}. This is a greater challenge for our model compared to CGE\cite{cge} since we tailor it for the open world.
Finally, we learn the mappings $\phi$ and $\omega$ by minimizing the cross-entropy loss over the cosine logits.
\begin{equation}
\label{eq:objective}
    \mathcal{L} = -\frac{1}{|\mathcal{T}|}\sum_{(x,c)\in\mathcal{T}} \log\frac{e^{\frac{1}{T}\cdot p(x,c)}}{\sum_{y\in\mathcal{C}^s}e^{\frac{1}{T}\cdot p(x,y)}}
\end{equation}
where $T$ is a temperature value that scales the probabilities values for the cross-entropy loss \cite{zhang2019adacos} and $p(x,c)=\cos(\phi(x),\omega(c))$. By exploiting graph embeddings and bounding the classifier scores for seen and unseen compositions, our \ours\ achieves outstanding performance on the closed world scenario. 
In the following we discuss how we extend \ours\ to the more challenging OW-CZSL.  

\subsection{From Closed to Open World CZSL}
\label{sec:compcos-open} 
OW-CZSL setting requires  
avoiding distractors, \ie unlikely concepts such as \textit{ripe dog}. 
The similarity among objects and states can be used as a proxy to estimate the feasibility of each composition. We can then inject the estimated feasibility into \ours\ both as margins in the loss function and as weights within the adjacency matrix.

\myparagraph{Estimating Compositional Feasibility.}
Let us consider two objects, namely \textit{cat} and \textit{dog}. We know, from our training set, that \textit{cats} can be \textit{small} and \textit{dogs} can be \textit{wet} since we have at least one image for each of these compositions. However, the training set may not contain images of \textit{wet cats} and \textit{small dogs}, which we know are feasible in reality. We conjecture that similar objects share similar states while dissimilar ones do not. Hence, it is safe to assume that the states of \textit{cats} can be transferred to \textit{dogs} and vice-versa. 

With this idea in mind, 
we define the feasibility score of s composition $c=(s,o)$ with respect to the object $o$ as:
\begin{equation}
\label{eq:obj}
\rho_{obj}(s,o)= \max_{\hat{o}\in\mathcal{O}^s} cos(\phi(o),\phi(\hat{o}))
\end{equation}
with $\mathcal{O}^s$ being the set of objects associated with state $s$ in the training set $\mathcal{C}^s$, \ie\ $\mathcal{O}^s =\{o | (s,o) \in \mathcal{C}^s\}$. Note, that the score is computed as the cosine similarity between the object embedding produced by the graph and the most similar other object with the target state, thus the score is bounded in $[-1,1]$. Training compositions get assigned the score of 1.
Similarly, we define the score with respect to the state $s$ as: 
\begin{equation}
\label{eq:attr}
\rho_{state}(s,o)= \max_{\hat{s}\in\mathcal{S}^o} cos(\phi(s),\phi(\hat{s}))
\end{equation}
with $\mathcal{S}^o$ being the set of states associated with the object $o$ in the training set $\mathcal{C}^s$, \ie\ $\mathcal{S}^o =\{s | (s,o) \in \mathcal{C}^s\}$. 
The feasibility score for a composition $c=(s,o)$ is then:
\begin{equation}
\label{eq:final-score}
\rho(c)=\rho(s,o)= g(\rho_{state}(s,o), \rho_{obj}(s,o))
\end{equation}
where $g$ is a mixing function, \eg\ max operation ($g(x,y)=\max(x,y)$) or the average ($g(x,y)=(x+y)/2$), keeping the feasibility score bounded in $[-1,1]$. Note that, while we focus on extracting feasibility from the visual information, 
external knowledge (\eg\ knowledge bases \cite{liu2004conceptnet}, language models \cite{wang2019language}) can be complementary resources.
A simple strategy to use the feasibility scores would be to consider all compositions above the threshold $\tau$ as valid and others as distractors: 
\begin{equation}
\label{eq:score-hard}
f_{\text{HARD}}(x) = \underset{c\in\mathcal{C}^t, \rho(c)>\tau}{\arg\max} \,\, cos(\omega(x),\phi(c)).
\myvspace{-3pt}
\end{equation}
However, this strategy might be too restrictive in practice. For instance, \textit{tomatoes} and \textit{dogs} being far in the embedding space does not mean that a state for \textit{dog}, \eg \textit{wet}, cannot be applied to a \textit{tomato}. Therefore, considering the feasibility scores as the golden standard may lead to excluding valid compositions (see Figure \ref{fig:method}).  
To sidestep this issue, we inject the feasibility scores directly into both the model and the training procedure. We argue that doing so enforces separation between most and least feasible unseen compositions in the shared embedding space. 

\myparagraph{Feasibility-aware objective.}
First, 
we integrate the feasibility scores $\rho(c)$  directly within our objective function as margins, defining the new objective as:
\begin{equation}
\label{eq:objective-feasibility}
    \mathcal{L}_{\text{f}} = -\frac{1}{|\mathcal{T}|}\sum_{(x,c)\in\mathcal{T}} \log\frac{e^{\frac{1}{T}\cdot p_\text{f}(x,c)}}{\sum_{y\in\mathcal{C}}  e^{\frac{1}{T}\cdot p_\text{f}(x,y)}}
    \end{equation}
with:
\begin{equation}
 \label{eq:margin-scores}
    p_\text{f}(x,c)= \begin{cases}
    \cos(\omega(x),\phi(c)) & \text{if}\,\, c\in\mathcal{C}^s\\
    \cos(\omega(x),\phi(c)) - \alpha\rho(c) &\text{otherwise}
    \end{cases}
\end{equation}
where $\rho(c)$ are used as margins for the cosine similarities, and $\alpha>0$ is a scalar factor.
With Eq.~\eqref{eq:objective-feasibility} we include the full compositional space while training with the seen compositions data to raise awareness of the margins between seen and unseen compositions directly during training. 

Note that, since $\rho(c_i)\neq\rho(c_j)$ if $c_i\neq c_j$ and $c_i,c_j \notin \mathcal{C}^s$, we have a different margin, \ie $-\alpha\rho(c)$, for each unseen composition $c$. 
In this way, we penalize less the more feasible compositions, pushing them closer to the seen ones, to which the visual embedding network is biased. 
At the same time, we force the network to push the representation of less feasible compositions away from the compositions in $\mathcal{C}^s$ in $\mathcal{Z}$. More feasible unseen compositions will then be more likely to be predicted by the model than the less feasible ones (which are more penalized).  As an example (Figure \ref{fig:method}, top part), 
the unfeasible composition \textit{ripe dog} is more penalized than the feasible \textit{wet tomato} during training, with the outcome that the optimization procedure does not force the model to reduce the region of \textit{wet tomato}, while reducing the one of \textit{ripe dog} (top-right pie).

We highlight that in this stage we do not explicitly bound the revised scores $p_\text{f}(c)$ to $[-1,1]$. Instead, we let the network implicitly adjust the cosine similarity scores during training. We also found it beneficial to linearly increase $\alpha$ till a maximum value as the training progresses, rather than keeping it fixed. This permits the model to gradually introduce the feasibility margins within the objective while exploiting improved primitive embeddings to compute them. 

\myparagraph{Feasibility-driven graph.}
Modelling the relationship between seen and unseen compositions via GCN is more challenging in the open world scenario, since also less feasible unseen compositions will influence the graph structure. This leads to two problems. The first is that distractors will influence the embeddings of seen compositions, making them less discriminative. The second is that the gradient flow will push unfeasible compositions close to seen ones, making harder to isolate distractors in the embedding space.

For these reasons, we modify the adjacency matrix of Eq.~\eqref{eq:cge-adj} in a weighted fashion, with the goal of 
reducing both the gradient flow on and the influence of less feasibile compositions. To achieve this goal, we directly exploit the feasibility scores, defining the adjacency matrix as:
\begin{equation}
    \label{eq:co-cge-adj}
    A_i^j =
        \begin{cases}
    \textcolor{black}{\max(0,\rho(k_i,k_j))}& \textcolor{black}{\text{if} \, (k_i,k_j)\in\mathcal{C},}\\
     \textcolor{black}{\max(0,\rho(k_j,k_i))}& \textcolor{black}{\text{if} \, (k_j,k_i)\in\mathcal{C},}\\
    1 & \text{if} \,k_j \in k_i\,\wedge\, k_i \,\in\mathcal{C},\\
    \max(0,\rho(k_i)) & \text{if} \,k_i \in k_j\,\wedge \, k_j \in\mathcal{C},\\
    1 & \text{if} \, i=j,\\
    0 &\text{otherwise.}\\
    \end{cases}
\end{equation}
In Eq.~\eqref{eq:co-cge-adj}, the connection between a state $s$ and an object $o$ corresponds to the feasibility of the composition $(s,o)$, such that the higher is the feasibility of the composition and the stronger is the connection among the two constituent primitives. Similarly, the influence of a composition to its primitives (third row) corresponds to the feasibility of the composition itself. 
We found that it is beneficial 
to influence the embedding of a composition $c=(s,o)$ fully by the embeddings of its primitives $s$ and $o$ (Eq.~\eqref{eq:co-cge-adj}, second row).
The motivation is that the mapping between compositions and primitives is not bijective: one composition corresponds to only one state and one object, but states and objects build multiple compositions. So while a composition is surely connected with its constituent primitives (second row, value 1), a state and an object are more related to existing, feasible compositions (third row). 

The formulation of Eq.~\eqref{eq:co-cge-adj} makes the connections in the graph 
dependent on the feasibility of the compositions. This allows the model to reduce the impact of less feasible compositions both in the forward pass and in the backward, making the shared embedding space more discriminative and less-influenced by distractors. 

\myparagraph{Discussion.} Our \ours\ model uses a GCN  
to map compositions to the shared embedding space, and a cosine classifier to measure the compatibility between image features and composition embeddings. This formulation merges and extends our previous models CGE \cite{cge} and CompCos \cite{compcos}. In particular, as in CGE we model the relationship between seen and unseen compositions through a graph. This allows us to perform end-to-end training of the CNN backbone without overfitting, since the feature representation is regularized by the compositional graph. 

Na\"ively applying CGE is not effective in the open world scenario, where we need to model the feasibility of each composition. Thus, following CompCos, we 
estimate the feasibility scores of each compositions and using the scores as margins in the objective function, with a cosine similarity-based classifier. 
We improve CompCos by modeling the feasibility of each composition also within the model by defining a weighted adjacency matrix for the GCN, with the weights dependent on the feasibility scores. Moreover, the primitive embeddings used to compute the feasibility scores, are produced by the GCN (thus influenced by the respective compositions) rather than learned in isolation, as in CompCos.
These modifications allow \ours\ to build a more discriminative shared embedding space where the compatibility function better isolates less feasible compositions. Finally, 
since the model is already robust enough to the presence of distractors in OW-CZSL, we do not need to use hard masking in Eq.~\eqref{eq:score-hard}.

%% file: sections/experiments.tex
{
\setlength{\tabcolsep}{2pt}
\renewcommand{\arraystretch}{1.4}
\begin{table}[t]
    \centering
    \resizebox{\linewidth}{!}
    {\begin{tabular}{l cc|cc|ccc|ccc}
         &   &  & \multicolumn{2}{c}{\textbf{Training}} & \multicolumn{3}{c}{\textbf{Validation}} & \multicolumn{3}{c}{\textbf{Test}}\\
        \textbf{Dataset}& s & o & sp  & i & sp  & up  & i & sp  & up  & i \\ 
    \hline
    MIT-States\cite{mitstates}     &  115 & 245 & 1262 & 30k & 300 & 300 & 10k & 400 & 400 & 13k \\
    UT-Zappos\cite{utzappos1} & 16 & 12 & 83 & 23k & 15 & 15 & 3k & 18 & 18 & 3k \\
    C-GQA (Ours) 
    & 413 & 674 & 5592 & 27k & 1252 & 1040 & 7k & 888 & 923 & 5k \\
    \end{tabular}}
    \caption{\textbf{Dataset statistics for CZSL}: We use three datasets to benchmark our method against the baselines. C-GQA (ours): our proposed dataset splits from Stanford GQA dataset \cite{gqa}. (s: \# states, o: \# objects, sp: \# seen compositions, up: \# unseen compositions, i: \# images)}
    \label{tab:dataset}
\end{table}
}

{
\setlength{\tabcolsep}{6pt}
\renewcommand{\arraystretch}{1.1}
\begin{table*}[t]
    \centering
    \resizebox{\linewidth}{!}{\begin{tabular}{ c  l| c c c c | c c c c | c c c c }
    \multirow{2}{*}{\color{black}\textbf{Training}} & \multirow{2}{*}{\textbf{Method}}  & \multicolumn{4}{c}{\textbf{MIT-States}}& \multicolumn{4}{c}{\textbf{UT-Zappos}}& \multicolumn{4}{c}{\textbf{C-GQA}}\\
                                                &   
                                                &S   & U    & HM    & AUC   
                                                &S   & U   & HM    & AUC   
                                                &S   & U    & HM    & AUC \\\hline
    \multirow{9}{*}{\STAB{\rotatebox[origin=c]{90}{\textbf{Closed}}}}& AoP\cite{aopp}    
    &14.3   &17.4       &9.9    &1.6 
     &{59.8}   &54.2       &40.8    &25.9
     & 17.0 & 5.6      & 5.9   &0.7\\
     &LE+\cite{redwine}       
     &15.0   &20.1       &10.7   &2.0   
     &53.0   &{61.9}       &41.0   &25.7&
     18.1 &  5.6     &    6.1&0.8\\
    &TMN\cite{tmn}      
    &20.2   &20.1       &13.0   &2.9      
     &58.7   &60.0       &45.0  &{29.3}
     & 23.1 & 6.5      & 7.5   & 1.1\\
    & SymNet\cite{symnet}          
    &24.2   &{25.2}       &{16.1}   &3.0 
    &49.8   &{57.4}       &{40.4}   &23.4 
     &26.8  &    10.3   & 11.0   &2.1\\
     &CompCos$^\text{CW}$    
     &{25.3}   &24.6       &{16.4}   &{4.5}   
     &{59.8}   &{62.5}       &{43.1}   & 28.1
     & 28.1 &    11.2   &    12.4& 2.6\\
    &CGE$_\text{ff}$    
    & 28.7 & 25.3      & 17.2   & 5.1&  
     56.8 & 63.6 & 41.2   & 26.4 
     & 28.1 &    10.1   &    11.4& 2.3\\
    &\ours$^\text{CW}_\text{ff}$
    & 27.8 & 25.2 & 17.5&5.1
     & 58.2 &63.3       &  44.1  &29.1 
     & 29.3 &    11.9   &12.7 & 2.8\\ 
   & CGE 
   & \textbf{32.8} & 28.0      &\textbf{21.4 }   &6.5 
     & \textbf{64.5} & \textbf{71.5} &  \textbf{60.5}  &  33.5
     & \textbf{33.5} &    \textbf{15.5}  &    \textbf{16.0}& \textbf{4.2}\\
        &  \ours$^{\text{CW}}$      
        & 32.1 & \textbf{28.3}      & 20.0   & \textbf{6.6}
     & 62.3 &    66.3   &    48.1&  \textbf{33.9}
     & {33.3}  & {14.9}      &   {15.5} &{4.1}\\\hline
    \multirow{3}{*}{\STAB{\rotatebox[origin=c]{90}{\textbf{Open}}}}&CompCos       
    &{25.6}   &22.7       &{15.6}   &{4.1} 
    &{59.3}   &{61.5}       &{40.7}   &{27.1}  
     & 28.4 & 10.7      &11.5    &2.4\\
    &\ours$_\text{ff}$     
    & 26.4  &  23.3     &  16.1  & 4.3
     & 60.1 &62.1       &  44.0 & 29.2 
     & 28.7 &  10.0 &  10.7  & 2.2\\
    &\ours 
    &  \textbf{30.3}     &  \textbf{24.1} & \textbf{17.3}   & \textbf{5.1}
     & \textbf{61.2} & \textbf{64.8}      &   \textbf{47.2} & \textbf{31.9}  
     &   \textbf{31.0}&  \textbf{13.3}     &   \textbf{14.1} & \textbf{3.4}\\
    \end{tabular}}
    \vspace{1pt}
    \caption{\textbf{Closed World CZSL results} on MIT-States, UT-Zappos and C-GQA. We measure 
    best seen (S) and unseen accuracy (U), best harmonic mean (HM), and area under the curve (AUC) on the compositions. {CW denotes closed-world version of the model, ff denotes frozen feature extractor.}}
    \myvspace{-10pt}
    \label{tab:sota-closed}
\end{table*}
}

\myparagraph{Datasets.}
We perform our experiments on three datasets (see Table \ref{tab:dataset}).
We adopt 
the standard split of MIT-States\cite{mitstates} from~\cite{tmn}. 
For the open world scenario, 
26114 out of 28175 ($\sim$93\%) are not present in any splits of the dataset but are included in our open world setting.
In UT-Zappos \cite{utzappos1,utzappos2}
we follow the splits from \cite{tmn}. 
Note that although 76 out of 192 possible compositions ($\sim$40\%) are not in any of the splits of the dataset, we consider them in our open world setting.

Both UT-Zappos and MIT-States have limitations. UT-Zappos\cite{utzappos1, utzappos2} is arguably not entirely compositional as states like \textit{Faux leather} vs \textit{Leather} are material differences not always observable as visual transformations. MIT-States instead contains images collected through older search engine with limited human annotation leading to significant label noise \cite{causal}. 
To address the limitations of these two datasets, in our previous work \cite{cge} we introduced a split 
built on top of Stanford GQA dataset \cite{gqa}, \ie  
the Compositional GQA (C-GQA) dataset. In this work we extend it to the OW-CZSL task. 
{With 413 states and 674 objects, the resulting OW-CZSL search space has almost 280K compositions making it way more challenging than other benchmarks.}  

\myparagraph{Metrics.} 
In zero-shot learning, models being trained only on seen $\mathcal{Y}_s$ labels (compositions) causes an inherent bias against the unseen $\mathcal{Y}_n$ labels. As pointed out by \cite{chao2016empirical, tmn}, the model thus needs to be calibrated by adding a scalar bias to the activations of the novel compositions to find the best operating point and evaluate the GCZSL performance. 

We adopt the evaluation protocol of \cite{tmn} and report the Area Under the Curve (AUC)~(in $\%$) between the accuracy on seen and unseen compositions at different operating points with respect to the bias. The best unseen accuracy is calculated when the bias term is large, \ie the model predicts only the unseen labels, also known as zero-shot performance. In addition, the best seen performance is calculated when the bias term is negative, \ie the model predicts only the seen labels. As a balance between the two, we also report the best harmonic mean (HM). 
We emphasize that our C-GQA dataset splits and the MIT-States and UT-Zappos dataset splits from \cite{tmn} do no not violate the zero-shot assumption as results are ablated on the validation set.
We therefore advice future works to also use our splits.

\myparagraph{Benchmark and Implementation Details.}
Following \cite{tmn,symnet} we use a ResNet18 pretrained on ImageNet~\cite{deng2009imagenet} as feature extractor $\omega$ and fine-tune the whole architecture with a learning rate of $5\cdot10^{-6}$, 
\ie \ours. For a fair comparison with the models that use a fixed feature extractor, we also perform experiments with a simplification of our model where 
we learn a 3 layer fully-connected (FC) network with ReLU\cite{nair2010rectified}, LayerNorm\cite{ba2016layer} and Dropout\cite{srivastava2014dropout} while keeping the feature extractor fixed, \ie  \ours$_\text{ff}$. 

We initialize the embedding function $\varphi$ with 300-dimensional {word2vec}~\cite{word2vec} embeddings for UT-Zappos and C-GQA, and with 600-dimensional {word2vec+fastext}~\cite{fasttext} embeddings for MIT-States, following \cite{xian2019semantic}, keeping the same dimensions for the shared embedding space $\mathcal{Z}$. We train both $\omega$ and $\phi$ using Adam~\cite{kingma2014adam} optimizer with a learning rate and a weight decay set to $5\cdot10^{-5}$. For both \ours\ and CompCos, the margin factor $\alpha$ and the temperature $T$ are set to $0.4$ and $0.05$ respectively for MIT-States, $1.0$ and $0.02$ for UT-Zappos, and $0.1$ and $0.02$ for C-GQA. We linearly increase $\alpha$ from 0 to these values during training,  reaching them after 15 epochs. We consider the mixing function $g$ as the average to merge state and object feasibility scores for both our model and CompCos. For CompCos we additionally use $f_\text{HARD}$ as predictor, unless otherwise stated. 

For $\phi$ we use a shallow 2-layer GCN with a hidden dimension of $4096$ in the closed world experiments, but for UT-Zappos, where we use a dimension of $300$. For the OW-CZSL experiments, we found beneficial to reduce the hidden dimension to the same of the input embeddings, i.e to $300$ for UT-Zappos and C-GQA, $600$ for MIT-States. Note that since the C-GQA search space is extremely high in the OW-CZSL setting, to test CGE and the closed world version of \ours, we reduce their hidden dimension to $1024$. {In the tables, we denote with the superscript CW the closed-world version of the models while with the subscript ff the method with frozen feature extractor.} 

We compare with four state-of-the-art methods, Attribute as Operators (AOP) \cite{aopp}, considering objects as vectors and states as matrices modifying them \cite{aopp}; LabelEmbed+ (LE+) \cite{redwine,aopp} training a classifier merging state and object embeddings with an MLP; Task-Modular Neural Networks (TMN) \cite{tmn}, modifying the classifier through a gating function receiving as input the queried state-object composition; and SymNet \cite{symnet}, learning object embeddings showing symmetry under different state-based transformations. We also compare \ours\ with our previous works, CGE \cite{cge} and CompCos \cite{compcos}. {Note that both CGE and Co-CGE$^{CW}$ need to produce embeddings for all unseen compositions to be applicable at inference time, thus, in OW-CZSL objects are connected to all states and with the compositions they take part in.}
We train each model with their default hyperparameters, reporting the closed and open world results of the models with the best AUC on the validation set. 
We implement our method in PyTorch\cite{paszke2019pytorch} and train on a Nvidia V100 GPU. For baseline comparisons, we use the authors' implementations where available.

\subsection{Closed World CZSL}
 \label{sec:exp-cw}
\myparagraph{Comparison with the State of the Art.}
We experiment with the closed world setting on the test sets of all three datasets.
Table \ref{tab:sota-closed} (top) shows models trained with the closed world assumption, while Table \ref{tab:sota-closed} (bottom) shows the open world models, not using any prior on the unseen test compositions during training but still predicting over a closed set. 
 
\oursCwEe\ achieves either comparable or superior results to the state of the art in all settings and metrics. In general, the results of \oursCwEe\ are comparable or superior to CGE, while surpassing by a margin the other approaches in the closed world, \eg on MIT-States \oursCwEe\ vs CGE achieves an AUC of 6.6 vs 6.5. This signifies the importance of graph based methods for CZSL as both outperform the closest non graph baseline CompCos by a large margin. {Note that the main difference between CGE and Co-CGE$^\text{CW}$ is the classifier, using cosine-similarity scores in Co-CGE$^\text{CW}$ while a linear operation in CGE. For this reason, we do not expect their performance to substantially differ in the closed-world setup.}

Similar observations apply to UT-Zappos, where \oursCwEe\ is superior to all methods in AUC (33.9 vs 33.5 of CGE). However CGE outperforms our model for the best seen, unseen and HM. This signifies that while our model does not achieve the best accuracies, it is less biased between the seen and unseen classes leading to a better AUC. {Our open-world model (Co-CGE) achieves lower results than the CGE counterpart (\eg 33.5 AUC of CGE vs 31.9 of Co-CGE). However, \ours\ is not using any prior on unseen classes during training. A model exploiting this prior can produce embeddings more discriminative for seen compositions, not influenced by non-existing unseen ones. Under this light our results are remarkable, since \ours\ achieves results close to the state of the art. Note that, the impact of such prior becomes more evident with end-to-end training, since a large number of parameters can exploit it. Without end-to-end training, our open world model (Co-CGE$_{\text{ff}}$) even surpasses CGE$_{\text{ff}}$, (\eg 29.2 vs 26.4 AUC).}

{Finally, in the challenging C-GQA dataset, \oursCwEe~achieves results comparable to CGE in terms of AUC (4.1 vs 4.2), almost twice the AUC of  SymNet (2.1) and almost four times the ones of TMN (1.1). Furthermore, \oursCw\ is the best among the non fine-tuned methods in all metrics (\eg w.r.t. CompCos$^\text{CW}$, 2.8 AUC vs 2.6, 12.7 HM vs 12.4, 11.9 unseen accuracy vs 11.2). Remarkably, \ours\ achieves better results than any closed-world method but CGE and \oursCwEe\ under all metrics (\ie 3.4 AUC, 14.1 HM and 13.3 unseen accuracy), despite not exploiting any prior on unseen compositions during training.}

{
\setlength{\tabcolsep}{6pt}
\renewcommand{\arraystretch}{1.2}
\begin{table}[t]
    \centering
    \resizebox{\linewidth}{!}
    {\begin{tabular}{l|cc}
\textbf{Connections in Graph} & \textbf{AUC} & \textbf{Best HM}   \\
\hline
{a) Visual Product} & {3.8} & {14.5}\\
b) Direct Word Embedding {average}& 5.9 & 19.4 \\
{c) Direct Word Embedding concat}& {6.1} & {19.7}\\
d) c$\rightarrow$ p, p$\rightarrow$ c, no self-loop on y  & 7.6 & 18.6 \\
e) c$\rightarrow$ p, p$\rightarrow$ c & 8.1 & 22.7 \\
f) {\textit{CGE}:} c$\rightarrow$ p, p$\rightarrow$ c, and s$\leftrightarrow$ o  & \textbf{8.6} & \textbf{23.3} \\
g) \textbf{\textit{\ours$^\text{CW}$}:} c$\rightarrow$ p, p$\rightarrow$ c, and s$\leftrightarrow$ o    & 7.9 & 22.5 \\
\end{tabular}}
\caption{\textbf{Ablation over the graph connections} validates the structure of our proposed graph on the validation set of MIT-States dataset. We start from directly using the word embeddings as classifier weights to learning a globally consistent embedding from a GCN as the classifier weights (s: states, o: objects, p: primitives, y: compositional labels). }
\label{tab:gcnconnection}
\end{table}
}

\myparagraph{Ablation Study.} We perform an ablation study with respect to the various connections in our compositional graph on the validation set of MIT-States and report results in Table \ref{tab:gcnconnection}. We start with standard cross-entropy loss, as in CGE and we then include the cosine similarity-based classifier. 

{As first baselines we consider three approaches not modeling the relationship between state-objects and their compositions. These approaches are \textit{Visual Product} (row a), classifying objects and states independently with two different predictors, and \textit{Direct Word Embedding average} (b) and \textit{Direct Word Embedding concat} (c), where the embeddings of each composition are initialized as the average (or concatenation) of its constituent object and state embeddings. As the results show, Visual Product achieves only 3.8 AUC and 19.5\% of best HM. Direct word embeddings achieve better results, with 5.9 AUC and 19.5\% HM when averaging, and 6.1 AUC and 19.7\% HM when concatenating them.}
{Interestingly, using cross-entropy as a loss function (rather than standard triplet or binary cross-entropy ones) provides already very good results w.r.t. the competitors (\eg 6.1 AUC of direct word embedding concat vs 4.3 reported in \cite{symnet}). This because cross-entropy directly models the relative similarity between all seen compositions and the ground-truth, opposite to the binary relationships of BCE and of sampled comparisons of the triplet loss. }

{Including the graph by simply connecting the primitives (\ie states and objects, p) to compositional labels (y) but without self connections for the compositional label (row d) achieves an improvement on AUC (7.6) while a slight decrease on best HM (18.6\%). 
 When we include self connections in the graph for compositions, (row e) outperform all approaches treating objects and states independently by a margin, with an AUC of 8.1 and a best HM of 22.7. This demonstrates the benefit of connecting objects, states and compositions.} 
Row (f) is the final CGE model, which additionally incorporates the connections between states (s) and objects (o) in a pair to model the dependency between them. We observe that learning a representation that is consistent with states, objects and the compositional labels increases the AUC from 8.1 to 8.6 validating the choice of connections in the graph. 
Finally, if we employ a cosine classifier to replace the dot product classifier of CGE, we see in row (g) that the AUC and HM are comparable. Note that, with this variant we can use the feasibility scores as margins in the objective tailoring the method for OW-CZSL. 

\subsection{Open World CZSL}
\label{sec:exp-ow}

{
\setlength{\tabcolsep}{6pt}
\renewcommand{\arraystretch}{1.1}
\begin{table*}[t]
    \centering
    \resizebox{\linewidth}{!}{\begin{tabular}{c  l| c c c c | c c c c |  c c c c }
    \multirow{2}{*}{\color{black}\textbf{Training}} & \multirow{2}{*}{\textbf{Method}} & \multicolumn{4}{c}{\textbf{MIT-States}}& \multicolumn{4}{c}{\textbf{UT-Zappos}}& \multicolumn{4}{c}{\textbf{C-GQA}}\\
                                                &
                                                &S   & U    & HM    & AUC  
                                                &S   & U   & HM    & AUC
                                                &S   & U    & HM    & AUC \\\hline
    \multirow{9}{*}{\STAB{\rotatebox[origin=c]{90}{\textbf{Closed}}}}  &AoP\cite{aopp}
     &16.6   &5.7       &4.7    &0.7 
     &  50.9 &  34.2     & 29.4   & 13.7
     & NA &  NA     &    NA&NA\\
      &LE+\cite{redwine} 
      &14.2   &2.5       &2.7   &0.3  
     & {60.4}  &   36.5    & 30.5  & 16.3 
     & 19.2 & 0.7 & 1.0  &0.08\\
      &TMN\cite{tmn}
      &  12.6        &  0.9     &   1.2 &   0.1  
     &     55.9  &18.1           & 21.7       &  8.4
     & NA  &    NA   & NA   & NA\\
      &SymNet\cite{symnet}  
      &21.4   &7.0       &5.8   &0.8              
     &53.3   &44.6       &34.5   &18.5
     &  26.7     &  2.2      &3.3    &0.43\\
      &CompCos$^\text{CW}$ 
      &25.3   &5.5       &5.9   &0.9 
    &59.8   &45.6   &36.3   &20.8
    & 28.0 &    1.0   &    1.6&0.20\\
     &CGE$_\text{ff}$ 
     & 29.6 & 4.0 & 4.9 & 0.7  
    & 58.8 & 46.5 & 38.0 & 21.5  
     &  28.3& 1.3      &    2.2&0.30\\
     &{\ours$^\text{CW}_\text{ff}$}    
     & 28.2 & 6.0 & 6.5 & 1.1  & 
    59.5 & 41.5 & 36.1  & 20.1
     & 28.9 &    1.2   & 2.1   & 0.29\\
     &CGE 
     &  \textbf{32.4}& 5.1      &   6.0 & 1.0
     & 61.7      & \textbf{47.7}  & 39.0      &  23.1  
      & \textbf{32.7} &   1.8    &   2.9 &0.47\\
       &    {\ours$^{\text{CW}}$} 
       & 31.1 &   5.8    &  6.4  & 1.1
     & \textbf{62.0}      & 44.3  & 40.3      &  23.1  
    & {32.1} & 2.0  &3.4    &0.53\\\hline
     \multirow{3}{*}{\STAB{\rotatebox[origin=c]{90}{\textbf{Open}}}}&{CompCos}   
     & {25.4}   &{10.0}       &{8.9}   &{1.6} & 
     59.3 & {46.8} & {36.9} & {21.3}
     &28.4  &      1.8 &    2.8&0.39\\
     &{\ours$_\text{ff}$} 
     & 26.4 &  10.4     &  10.1  & 2.0
     &60.1 &    44.3   & 38.1   &21.3 
     & 28.7 &    1.6   &    2.6&0.37\\
     &{\ours} 
     & 30.3  & \textbf{11.2} & \textbf{10.7}& \textbf{2.3}
     & 61.2 &      45.8 &    \textbf{40.8}& \textbf{23.3}
     & 32.1 &     \textbf{3.0}  & \textbf{4.8}&\textbf{0.78}\\
    \end{tabular}}
    \vspace{1pt}
    \caption{\textbf{Open World CZSL results} on MIT-States, UT-Zappos and C-GQA. We measure best seen (S) and unseen accuracy (U), best harmonic mean (HM), and area under the curve (AUC) on the compositions. {CW denotes closed-world version of the model, ff denotes frozen feature extractor.}}
    \myvspace{-8pt}
    \label{tab:sota-open}
\end{table*}
}

\myparagraph{Comparing with the State of the Art.}
As shown in Table \ref{tab:sota-open}, the first clear effect of moving to the more challenging OW-CZSL setting is the severe decrease in performance for every method. The largest decrease in performance is on the best unseen metric, due to the presence of a large number of distractors. As an example, in MIT states LE+ goes from 20.1\% to 2.5\% of best unseen accuracy and even the previous state of the art, CGE, loses 22.9\%. Similarly, in C-GQA the best seen accuracy drops of 6.8$\%$ for SymNet, $8.7\%$ for CompCos$^\text{CW}$ and even the end-to-end trained CGE and \oursCwEe\ lose 11.8$\%$ and $12.3\%$ respectively. 

Compared to the baselines, our models, \ours$_\text{ff}$\ and \oursEe\ are more robust to the presence of distractors, \eg 
particularly for the best HM performance on MIT-States, \ours$_\text{ff}$\ surpasses CompCos by 1.2$\%$ at the same feature extractor. 
This demonstrates the importance of explicitly modeling the different feasibility of the compositions in the whole compositional space, injecting them within the objective function and the graph connections. Similar considerations apply to \oursEe, that achieves the best results in MIT-States and C-GQA wrt all metrics. Remarkably, it achieves a 0.78 of AUC on C-GQA which is comparable to the closed world results of early CZSL methods, such as AoP (0.7) and LE+ (0.8). Over CompCos, the improvements are clear also in the accuracy on unseen classes (+1.8$\%$ on MIT-States, +1.2 $\%$ on C-GQA) and  harmonic mean (+1.8$\%$ on MIT-States, and $+2.0\%$ on C-GQA). {Interestingly, MIT states contains label noise, mostly linked to ambiguities in attribute annotations \cite{causal}. Under this light, the performance of \ours\ are remarkable, showing also robustness to label noise even in OW-CZSL. 
}
In UT-Zappos the performance gap with the other approaches is more nuanced. This is because the vast majority of compositions in UTZappos are feasible, thus it is hard to see a clear gain {from injecting} the feasibility scores into the training procedure. Nevertheless, \oursEe\ achieves the best HM mean (40.8$\%$) and AUC (23.3). However, a good closed world model performs well in this scenario, as showed by the performance of CGE, achieving the best accuracy on unseen classes (47.7$\%$). 
However, the overall results being lower than the closed setting indicates that 
OW-CZSL setting poses an open challenge. 

For C-GQA, we observe that due to the huge search space (almost 280k compositions) achieving good OW-CZSL performance is extremely hard compared to the much smaller search space of MIT-States (almost 30k compositions) and UT-Zappos (192). {It is interesting to highlight how the relative performance degradation of Co-CGE in moving to the OW-CZSL is 27.0\% for UT-Zappos (31.9 vs 23.3 AUC), 54.9\% on MIT-States (5.1 vs 2.3 AUC), and 77.1\% on C-GQA (3.4 vs 0.78 AUC).  
These results confirm how the larger is the compositional space, the larger is the decrease in performance of CZSL models. } Finally, Table \ref{tab:sota-open} shows two interesting trends. {The first is the importance of end-to-end training, with CGE and \oursCwEe\ surpassing all other methods but \oursEe.} The second, is the results of SymNet being slightly superior than \ours$_\text{ff}$\ in AUC, \ie SymNet is the only method modeling states and objects separately at classification levels. Therefore, as the search spaces grows, it may beneficial to predict each primitive independently to get an initial estimate of the composition.

 \setlength{\tabcolsep}{3pt}
\renewcommand{\arraystretch}{1.1}
\begin{table}[t]
    \centering
    \resizebox{\linewidth}{!}{\begin{tabular}{l l | c c c| c c c c}
      &  & & & & S & U    & HM    & AUC\\\hline
     \parbox[t]{2mm}{\multirow{4}{*}{\rotatebox[origin=c]{90}{\bf Margins}}}   & CompCos$^\text{CW}$  & \multicolumn{3}{l|}{} & 28.0 &6.0 &7.0 &1.2\\
        & CompCos & \multicolumn{3}{l|}{$\alpha=0$}& 25.4 &10.0 &9.7 &1.7\\
        & &\multicolumn{3}{l|}{+ $\alpha>0$} & 27.0 & 10.9& 10.5  &2.0 \\
        & &\multicolumn{3}{l|}{+ warmup $\alpha$}  & 27.1&{11.0}&10.8& 2.1\\
         \hline
       \parbox[t]{2mm}{\multirow{4}{*}{\rotatebox[origin=c]{90}{\bf Primitives}}} & \multirow{4}{*}{CompCos} &\multicolumn{3}{l|}{$\rho_{state}$} & 26.6 & 10.2& 10.2&1.9\\
       &  &  \multicolumn{3}{l|}{$\rho_{obj}$}  &27.2& 10.0 &9.9  &1.9\\
       & &\multicolumn{3}{l|}{$\max(\rho_{state},\rho_{obj})$} &27.2&10.1&10.1& 2.0  \\
        & &\multicolumn{3}{l|}{$(\rho_{state}+\rho_{obj})/2$}   &27.1&{11.0}&10.8& 2.1 \\\hline
               & & s$\leftrightarrow$o&p$\rightarrow$ c&c$\rightarrow$ p&\\\cline{3-5}
       \parbox[t]{2mm}{\multirow{9}{*}{\rotatebox[origin=c]{90}{\bf Graph Connections}}} & \multirow{9}{*}{\ours$_\text{ff}$} 
        & 1& 1& $\rho$&28.4& 9.1 &  9.3 & 1.8\\
        && 1& $\rho$& $\rho$&{26.6}& 8.1 &  8.4 & 1.5\\
        && $\rho$& 1& $\rho$&28.2& 8.9 &  9.0 & 1.7\\
        && $\rho$& $\rho$& $\rho$&29.3& 9.7 &  10.0 & 2.0\\
        && 1&1&1&28.4& 11.3 &11.6  & 2.4\\
        && 1& $\rho$& 1&28.2& 10.8 &  11.0 & 2.3\\
        & & $\rho$& 1&1&27.2& 11.9 &  11.6 & 2.3\\
        && $\rho$& $\rho$& 1&29.5& 11.5 &  12.0 & 2.5\\\cline{2-9}
        &\ours & \multicolumn{3}{l|}{end-to-end}  &30.4 & 12.4& 12.6&2.8\\
        
    \end{tabular}}
    \vspace{1pt}
    \caption{Results on MIT-States validation set for different ways of applying the margins (top), computing the feasibility scores (middle) and using the scores within the graph (bottom) for CompCos and \ours$_\text{ff}$. (S: seen, U: unseen)}
    \myvspace{-16pt}
    \label{tab:ablation-feasibility}
\end{table}

In the following experiments, we use MIT-States' validation set to analyze the different choices for our feasibility scores. In particular, we investigate the impact of the feasibility-based margins within the loss functions (starting from CompCos), how they are computed, and how they should be injected within the graph connections. Finally we check the eventual benefit that limiting the output space during inference using $f_{\text{HARD}}$ may bring to different models. 

\myparagraph{Importance of the feasibility-based margins.}   
We check the impact of including all compositions in the objective function (without any margin) and of including the feasibility margin but without any warmup strategy for $\alpha$. 
As the results in Table \ref{tab:ablation-feasibility} (Top) show, including all unseen compositions in the cross-entropy loss without any margin (\ie $\alpha=0$) increases the best unseen accuracy by 4\% and the AUC by 0.5. This is a consequence of the training procedure: since we have no positive examples for unseen compositions, including unseen compositions during training makes the network push their representation far from seen ones in the shared embedding space. This strategy regularizes the model in presence of a large number of unseen compositions in the output space. Note that this problem is peculiar in the open world scenario since in the closed world the number of seen compositions is usually larger than the unseen ones. The CompCos ($\alpha=0$) model performs worse than CompCos$^\text{CW}$~ on seen compositions, as the loss treats all unseen compositions equally.

Results increase if we include the feasibility scores during training (\ie $\alpha>0$). The AUC goes from 1.7 to 2.0, with consistent improvements over the best seen and unseen accuracy. This is a direct consequence of using the feasibility to separate the unseen compositions from the unlikely ones. In particular, this brings a large improvement on S and moderate improvements on both U and HM.

Finally, linearly increasing $\alpha$ (\ie warmup $\alpha$) further improves the harmonic mean due to both the i) improved margins estimated from the updated primitive embeddings and ii) the gradual inclusion of these margins in the objective.  
This strategy {improves} the bias between seen and unseen classes (as for the {better} on harmonic mean) while slightly {enhancing} the discriminability on seen and unseen compositions in isolation.

\myparagraph{Effect of Primitives.} 
We can either use objects as in Eq.~\eqref{eq:obj}, states as in Eq.~\eqref{eq:attr}) or both  as in Eq.~\eqref{eq:final-score} to estimate the feasibility score for each unseen composition. Here we show the impact of these choices on the results in Table \ref{tab:ablation-feasibility} (Middle).

We observe that computing feasibility on the primitives alone is already beneficial (achieving an AUC of 1.9) since the dominant states like \textit{caramelized} and objects like \textit{dog} provide enough information to transfer knowledge. 
In particular, computing the scores starting from state information ($\rho_{\text{state}}$) brings improves the best U and HM. 
Using similarities among objects ($\rho_{\text{obj}}$) performs well on S while achieving slightly lower performances on U and HM.

Introducing both states and objects give the best result at AUC of 2.1 as it combines the best of both. Merging objects and states scores through their maximum ($\rho_{\text{max}}$) maintains the higher seen accuracy of the object-based scores, with a trade-off between the two on unseen compositions. However, merging objects and states scores through their average brings to the best performance overall, with a significant improvement on unseen compositions (almost 1\%) as well as the harmonic mean. 
This is because the model is less-prone to assign either too low or too high feasibility scores for the unseen compositions, smoothing their scores. 
As a consequence, more meaningful margins are used in Eq.~\eqref{eq:objective-feasibility} and thus the network achieves a better trade-off between discrimination capability on the seen compositions and better separating them from unseen compositions (and distractors) in the shared embedding space. 

\myparagraph{Effect of Graph Connections.} 
For each graph connection, we have two choices: either keeping it unaltered (\ie value 1) or replacing it with the feasibility scores ($\rho$), as in Eq.~\eqref{eq:co-cge-adj}. In Table \ref{tab:ablation-feasibility} (bottom), we analyze these choices, \ie symmetric state and objects connections (s$\leftrightarrow$o), the connection from primitives to compositions (p$\rightarrow$c) and viceversa (c$\rightarrow$p).

A clear observation is the importance of keeping the influence of primitives on compositions (c$\rightarrow$p) unaltered (equal 1). We conjecture that since most of unseen compositions will get feasibility scores lower than 1, their representations would be updated mainly through their self-connection. However, the seen compositions for which we have supervision fully exploit the representations of their primitives. This causes the representations for unseen compositions to have an inherent distribution shift and being i) poorer with respect to seen one and ii) less discriminative. This is clearly shown in the table from the low HM and best unseen class accuracy of \ours\ whenever c$\rightarrow$p is different than 1. Keeping the connections c$\rightarrow$p as 1 allows the model to keep its discrimination capability on unseen classes and a best trade-off between accuracy on seen and unseen classes, with an average improvement of 0.63 in AUC.

For the other two types of connections, (s$\leftrightarrow$o) and (p$\rightarrow$c), the best results are achieved when their weights are set as in Eq.~\eqref{eq:co-cge-adj}, using the feasibility scores. This allows the model to achieve the best AUC (2.5$\%$), HM (12$\%$) and seen accuracy (29.5$\%$) while being slightly inferior to the top method in best unseen accuracy (-0.4$\%$). Note that the advantage of this combination is consistent also for c$\rightarrow$p set through $\rho$.

As a final experiment, we check the benefits of fine-tuning the while representation end-to-end with the best performing combination (c$\rightarrow$p and c$\rightarrow$p through $\rho$, c$\rightarrow$p). As expected this brings to the best results for all metrics. In particular, the learned representations results more discriminative for both seen and unseen compositions, achieving an improvement of $0.9\%$ on both best seen and best unseen accuracies and a consequent gain of $0.6\%$ in HM. 

\begin{figure}[t]
         \centering
    \includegraphics[width=0.48\textwidth]{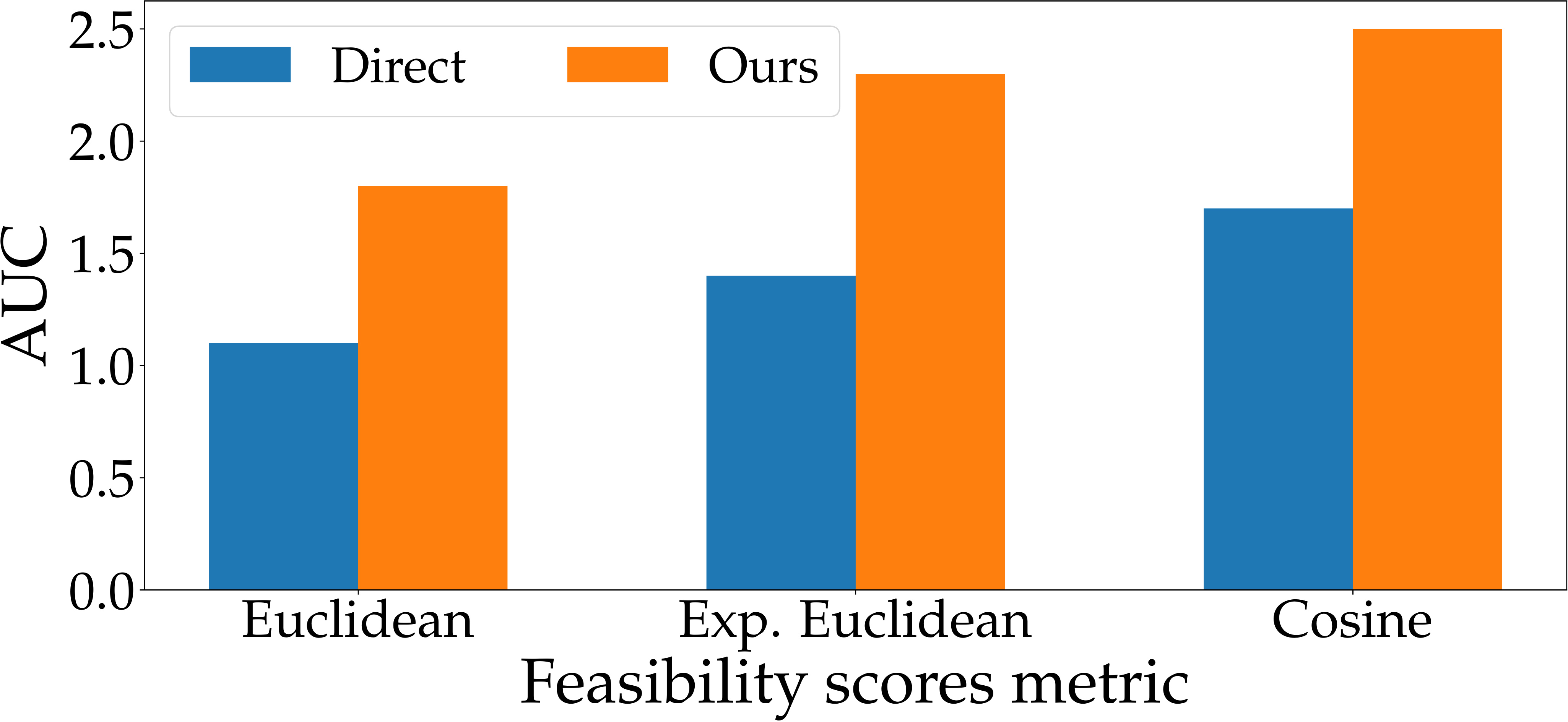}
         \centering\vspace{-5pt}
    \label{fig:}
    \caption{\color{black} AUC on MIT-States validation set by computing feasibility scores through different strategies: direct similarity between state and object embeddings (\textit{Direct}), \textit{Ours} (Section \ref{sec:compcos-open}), negative (\textit{Euclidean}) and exponential negative (\textit{Exp. Euclidean}) Euclidean distance, and cosine similarity (\textit{Cosine}).}
    \label{fig:euclidean-test}
\end{figure}

\setlength{\tabcolsep}{2pt}
\renewcommand{\arraystretch}{1.2}
\begin{table}[t]
    \centering
    \begin{tabular}{l | c | c c c c| l | c | c c c c}
       & Mask   & S & U    & HM    & AUC &  & Mask   & S & U    & HM    & AUC\\\hline
        \multirow{3}{*}{a.} & &\multirow{3}{*}{14.8} & 3.1&3.2 &0.3 & \multirow{3}{*}{d.}  & &  \multirow{3}{*}{28.0} &6.0 &7.0 &1.2\\
        & CompCos & &5.0& 4.6&0.5 & &  CompCos&  &  8.1&8.7& 1.6 \\
        & \oursEe& &6.1 &5.4&0.6& & \oursEe&&9.3&9.4 & 1.8 \\\hline
                  
        \multirow{3}{*}{b.}  & &\multirow{3}{*}{15.9} &1.3 &1.7 &0.1& \multirow{3}{*}{e.}  & &\multirow{3}{*}{31.2} &7.1 & 8.3 &1.6\\
         & CompCos& &4.1&4.1 &0.4 & &  CompCos && 8.5 & 9.5 &  2.0\\
        & \oursEe&& 4.7 &5.3&0.5 & & \oursEe&&11.1&11.5 & 2.6 \\\hline
                  
         \multirow{3}{*}{c.}  & &\multirow{3}{*}{23.6} & 7.9&7.6 &1.2 & \multirow{3}{*}{f.}  &  &\multirow{3}{*}{30.4} &12.4 & 12.6 &2.8\\
         & CompCos& &7.9& 7.7&1.2 &  &  CompCos && 12.5 & 12.6 &  2.8\\
        & \oursEe& & 8.7&8.4&1.4 & &\oursEe&&12.7&12.8 & 2.9 \\
    \end{tabular}
    \vspace{1pt}
    \caption{Results on MIT-States validation set by applying feasibility-based binary masks ($f_{\text{HARD}}$) on the methods a. LE+, b. TMN, c. SymNet,  d. CompCos$^\text{CW}$, e. \oursCwEe, f. \oursEe. (S: seen, U: Unseen)}
    \myvspace{-15pt}
    \label{tab:ablation-inference}
\end{table}

\begin{figure*}[t]
    \begin{subfigure}{0.48\textwidth}
         \centering
    \includegraphics[width=\textwidth]{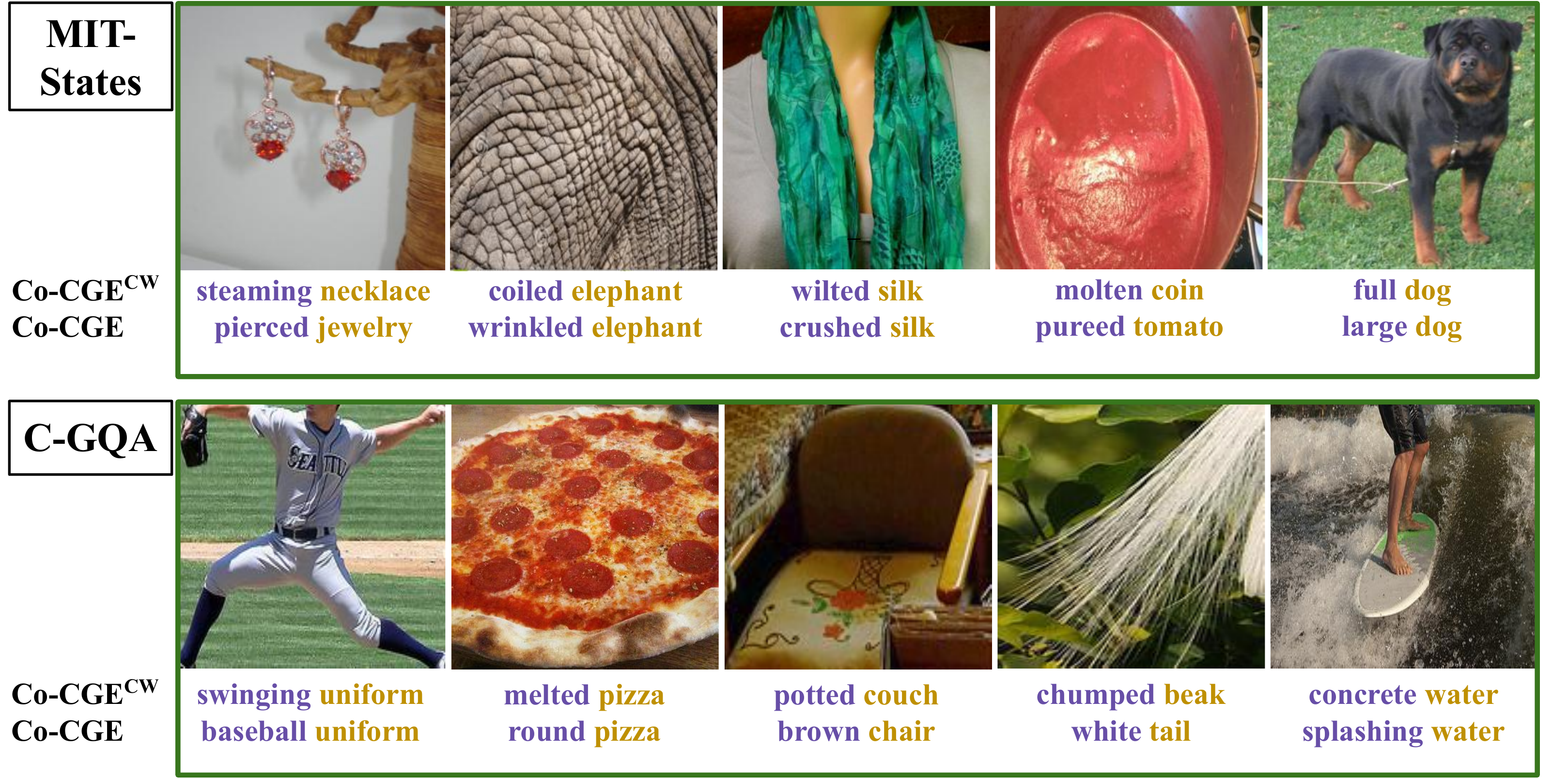}
         \centering\vspace{-10pt}
    \caption{Positive qualitative results.}
    \label{fig:qualitative-positive}
    \end{subfigure}
    \hspace{2mm}
    \begin{subfigure}{0.48\textwidth}
    \includegraphics[width=\textwidth]{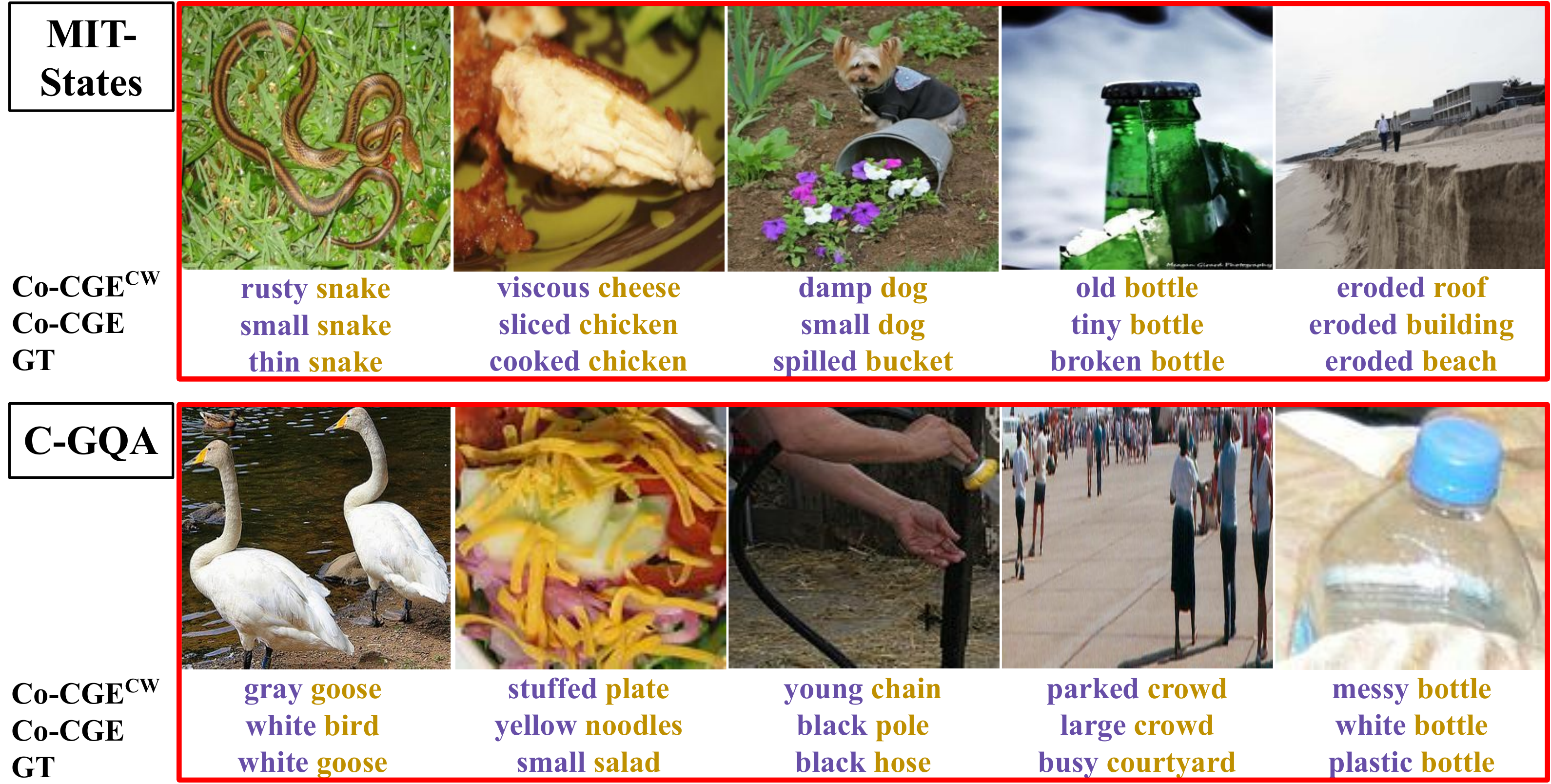}
         \centering\vspace{-10pt}
    \caption{\color{black} Failure cases.}
        \label{fig:qualitative-negative}
    \end{subfigure}
    \caption{\color{black} \textbf{Qualitative results.} Positive (left) and negative (right) examples of \oursEe\ predictions in the OW-CZSL scenario for MIT-States (top) and C-GQA (bottom), compared with \oursCwEe.}
    \vspace{-10pt}
    \label{fig:qualitative}
\end{figure*}

\begin{figure}[t]
         \centering
    \includegraphics[width=0.5\textwidth]{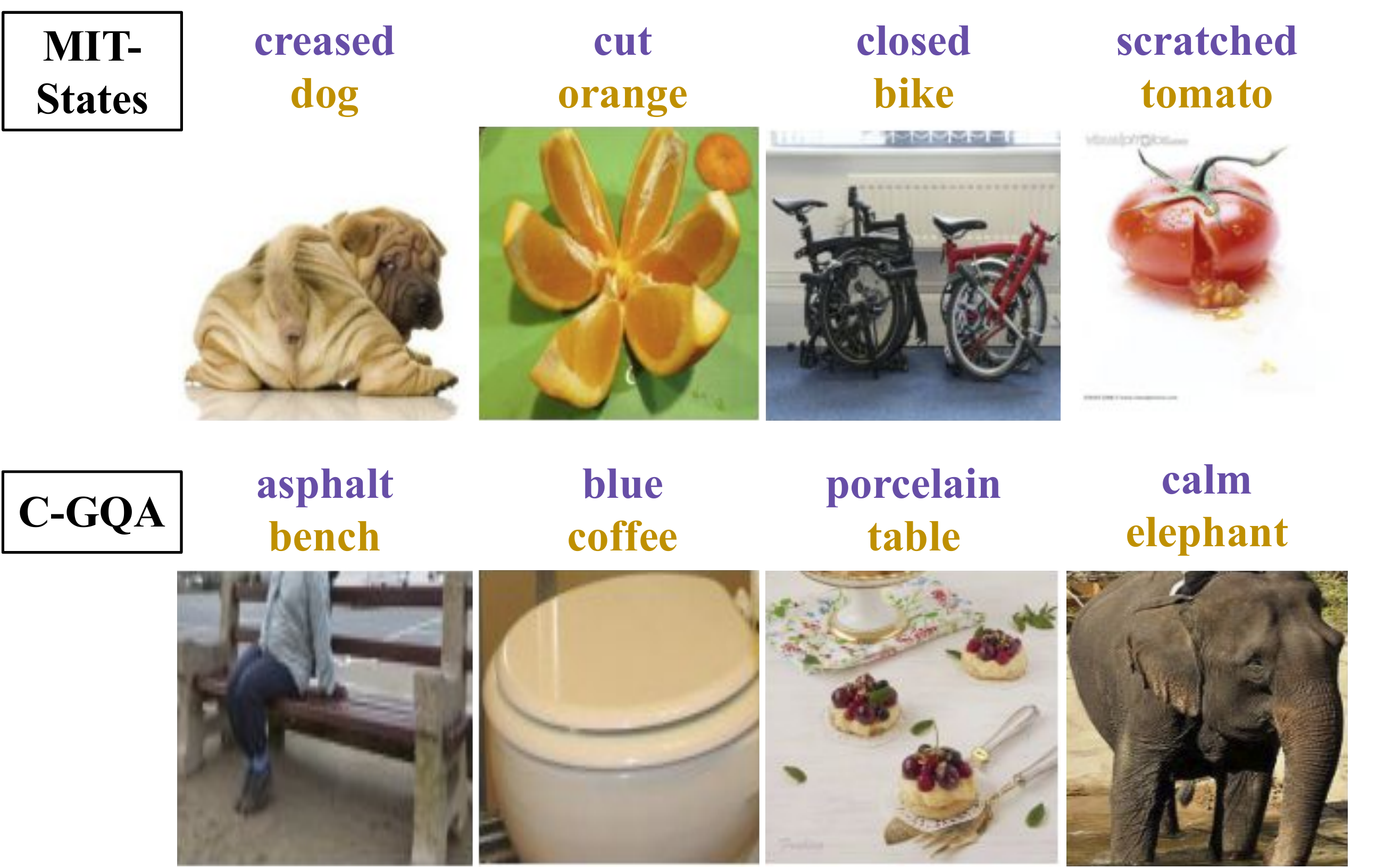}
    \vspace{-15pt}
    \caption{\textbf{Retrieving images from labels.} Top-1 retrieved images by \oursEE\ for state-object compositions not present in MIT-States (top) and C-GQA (bottom). } 
    \label{fig:retrieval}
\end{figure}

\myparagraph{\color{black}Effect of the Feasibility Computation strategy.}
{In Section \ref{sec:compcos-open}, we consider the cosine similarity to compare the primitive embeddings for computing the feasibility scores, propagating feasible states for an object using its similarity with other objects, and viceversa for the feasible objects for certain states. In this part, we explore alternatives to this choice. First, we test whether directly comparing object and state embeddings (\textit{Direct}) can outperform our strategy (Sec.~\ref{sec:compcos-open}). Second, we check if cosine similarity is the best similarity metric in this scenario, testing as alternative either negative Euclidean distance (Euclidean) or its negative exponential version (Exp. Euclidean), using in both cases the negative exponential of the distance to set the weights of the graph connections (to keep them positive). Note that, in the Euclidean case, the classifier is a non-bounded one (\ie linear) while in Exp. Euclidean we keep our cosine classifier (since scores are bounded in the range $[0,1]$).}

{Results are shown in Figure \ref{fig:euclidean-test} for the validation set of MIT-States. As the Table shows, our strategy (\textit{Ours}) largely outperforms the {Direct} counterpart for all similarity metrics, with an average improvement of 0.8 in AUC (\eg 1.7 vs 2.5 in AUC for \textit{Cosine}). We ascribe this behaviour to the different nature of objects and states: while objects are physical entities, states are modifiers modifying objects' appearance \cite{aopp} and the representation of a state is influenced by all objects it can be applied to. This leads to high feasibility scores for compositions with high correlation between objects and states (\eg \textit{inflated balloon}) while low feasibility scores whenever a state can be applied to a large variety of objects (\eg \textit{wet}). Our strategy sidesteps this problem by exploiting priors from training compositions and treating similarity between objects and states independently.
}

{For what concerns the similarity metrics, that cosine similarity achieves the best results, independently on the way feasibility scores are computed (\ie 1.7 AUC on {Direct} vs \textit{2.3} of {Exp. Euclidean} and 1.8 of {Euclidean}, {2.5} AUC on \textit{Ours} vs \textit{1.4} of {Exp. Euclidean} and 1.1 of {Euclidean}). These results show the importance of bounding the feasibility scores in a predefined range to: i) directly compare their values with the model's output, which is beneficial for the loss function, and ii) easily pruning graph connections by considering as feasible every composition with a score falling in the positive half of the range. 
Interestingly, also Exp. Euclidean achieves good results, confirming the importance of using bounded feasibility scores as margins in the loss function.}

\myparagraph{Effect of Hard Thresholding.} In Section \ref{sec:method}, we described that the feasibility scores can also be used during inference to mask the predictions, 
\ie using as prediction function Eq.~\eqref{eq:score-hard} in place of Eq.~\eqref{eq:prediction}, with the threshold $\tau$ computed empirically. 
We study the impact of this masking computed either with CompCos or with \oursEe\ feasibility scores on \oursEe\ and the closed world models  \ours$^\text{CW}$,  
LE+, TMN, SymNet and CompCos$^\text{CW}$, showing the results in Table \ref{tab:ablation-inference}. Note that, since seen compositions are not masked, best S performances do not change.

We observe that applying either ours or CompCos feasibility-based binary masks on top of all closed world approaches 
is beneficial. This is because the masks are able to filter out the less feasible compositions, rather than simply restricting the output space. In particular, CompCos-based mask brings an average improvement of 0.3 of AUC, 1.4\% of HM and 1.6\% of best unseen accuracy, while \oursEe-based improves the AUC of the base model of 0.5 in average, 2.4 in harmonic mean and 2.9 in best unseen accuracy. This suggests that \oursEe\ estimates more precise feasibility scores than CompCos, and it can better filter out compositions from the output space. 

Interestingly, masking largely improves \oursCwEe\ 
(+1.0 AUC), despite being the best performing closed-world method. On the other hand, SymNet does not benefit of CompCos-based masking (\eg +0.1 HM) and marginally does with \oursEe\ mask (\ie +0.8\% on unseen accuracy). 
This suggests that masking the output space with feasibility scores is more beneficial for models predicting object and states together at inference time (\eg LE+, \oursCwEe) than for those predicting them in isolation (\eg SymNet).
Finally, the improvements are minimal (\ie +0.1\% AUC) for our open world \oursEe\ model, being already robust  to the presence of distractors.  
Since  
$f_\text{HARD}$ requires tuning an additional hyperparameter, we do not apply any masking to \ours$_\text{ff}$ and \ours\ in the experiments. 

\input{sections/cross-dataset}

\subsection{Qualitative results} 
\label{sec:qualitatives}

\myparagraph{Influence of Feasibility Scores on Predictions.}
We analyze the reasons for the improvements of \oursEe\ over \oursCwEe, by showing output examples of both models on images of MIT states (top) and C-GQA (bottom). {In Figure \ref{fig:qualitative} we compare predictions on unseen compositions for samples where the closed world model is ``distracted'' while the open world one predicts the correct label (green, Fig.~\ref{fig:qualitative-positive}) and examples where also our model fails (red, Fig.~\ref{fig:qualitative-negative}).  

We observe that the closed world model is generally not capable of dealing with the presence of distractors. For instance, there are cases where the object prediction is correct (\eg \textit{coiled elephant}, \textit{wilted silk}, \textit{full dog}, \textit{rusty snake}, \textit{viscous cheese}, \textit{swinging uniform}, \textit{messy bottle}, \textit{concrete water}) but the associated state is not only wrong but also making the compositions unfeasible. In other cases, the state prediction is either correct (\eg \textit{eroded}) or reasonable (\eg \textit{melted}) but the associated object is not, eventually resulting in unfeasible predictions (\eg \textit{parked crowd} vs \textit{airplane} in the background). 
In some cases, both state and object predictions are incorrect for \oursCwEe, being either unfeasible (\eg \textit{steaming necklace}, \textit{young chain}, \textit{potted couch}) or confused in the large open world search space (\textit{molten coin}, \textit{old bottle}). All these problems are less severe in our full \ours\ model since injecting the feasibility of each composition within the objective and the graph connections helps in both reducing the possibility to predict implausible distractors and improving the structure of the compositional space,  better discriminating the constituent visual primitives. This occurs even when the predictions of \ours\ are wrong, being either close to the ground-truth (\ie \textit{small snake}, \textit{sliced chicken}) or referring to another concept of the image (\ie \textit{small dog}, \textit{black pole}).}

{Despite these results, the wrong predictions in Figure~\ref{fig:qualitative-negative} highlight some of the limitations of our current model. In particular, the model currently has no localization constraints on the object and state predictions. As a consequence, the predicted state might be linked to an object different than the predicted one (\eg \textit{eroded building}, \textit{white bottle}). Moreover, \oursEE\ does not model foreground information and might focus on different parts of the image than the target ones (\eg \textit{small dog}, \textit{black pole}, \textit{yellow noodles}). Furthermore, the focus on modeling the distractors might make the model less confident in discriminating known concepts, producing imprecise predictions (\eg \textit{white bird} vs \textit{white goose}). In the future it will be interesting to revisit the trade-off between discriminating seen compositions and isolating distractors as well as developing constraints to ensure that state and object predictions refer to the same part of the image. Finally, it will be crucial to break the limits of the current CZSL problem formulation, designing an approach that can predict multi-states/objects per image.}

\setlength{\tabcolsep}{2pt}
\renewcommand{\arraystretch}{1.2}
\begin{table}[t]
    \centering
    \resizebox{\linewidth}{!}{
    \begin{tabular}{ l  l  l }
\hline
        \textbf{MIT-States} & \multicolumn{2}{c}{\textbf{States}} \\
        & Most Feasible (Top-3)  &  Least Feasible (Bottom-3) \\
        \hline
         cat& huge, tiny, small & cloudy, browned, standing\\
         tomato &diced, peeled, mashed &full, fallen, heavy\\
         house & ancient, painted, grimy  &mashed, wilted, browned\\
         banana&diced, browned, fresh& dull, barren, unpainted\\
         knife&blunt, curved, wide&viscous, standing, runny\\

           \hline
        \textbf{C-GQA} & \multicolumn{2}{c}{\textbf{States}} \\
        & Most Feasible (Top-3)  &  Least Feasible (Bottom-3) \\
        \hline
         dog & small, drinking, toy &horizontal, ridged, styrofoam\\
         wine& pink, red, black & rubber, hard, angled\\
         fruit& unripe, sliced, peeled & angled, glossy, toilet\\
         jacket & tight, closed, sleeveless  &porcelain, shaped, miniature\\
         window&glass, beige, purple & packaged, greasy, boiled\\
    \end{tabular}
    }
    \vspace{1pt}
    \caption{ Top-3 highest and Bottom-3 lowest feasible state per object on MIT-States (top) and  C-GQA (bottom) as given by \oursEe.}
    \myvspace{-18pt}
    \label{tab:most-feasible-general}
\end{table}

\myparagraph{Discovering Most and Least Feasible Compositions.} The biggest advantage of our method is its ability to estimate the feasibility of each unseen composition, to later inject these estimates into the learning process and the model. Our procedure described in Section \ref{sec:compcos-open} needs to be robust enough to model which compositions should be more feasible in the {compositional} space and which should not, isolating the latter in the shared embedding space. We highlight that here we are focusing mainly on visual information to extract the relationships. This information can be coupled with knowledge bases (\eg \cite{liu2004conceptnet}) and language models (\eg \cite{wang2019language}) to further refine the scores.

Table \ref{tab:most-feasible-general} shows the top-3 most feasible compositions and bottom-3 least feasible compositions given five randomly selected objects for MIT-States (top) and C-GQA (bottom). These objects specific results show a tendency of the model to relate feasibility scores to the subgroups of classes. For instance, cooking states are considered as unfeasible for standard objects (\eg \textit{mashed house}, \textit{boiled window}) as well as substance-specific states (\eg \textit{runny knife}). Similarly, states usually associated with substances are considered unfeasible for animals (\eg \textit{runny cat}). At the same time, size and actions are mostly {linked with} animals (\eg \textit{small cat}, \textit{drinking dog}) while cooking states are correctly associated with food (\eg \textit{diced tomato}, \textit{sliced fruit}). 

Interestingly, in MIT-States the top states for \textit{knife} are all present with \textit{blade} as seen compositions, and in C-GQA the top states for \textit{dog} are all present with animals as seen compositions (\eg \textit{drinking cat}, \textit{small cat}, \textit{small horse}, \textit{toy cat}). This shows that our model exploits the similarities between two objects to transfer these states, \eg from \textit{blade} to \textit{knife} and from \textit{cat} to \textit{dog}, following Eq.~\eqref{eq:obj}. 
Furthermore, the state \textit{standing} is considered as unfeasible for \textit{cat} in MIT-States while being feasible (6th top) for \textit{dog} in C-GQA. This is because the state \textit{standing} has different meanings in the two datasets, \ie buildings (\eg \textit{standing tower}) in MIT-States, animals and persons (\eg \textit{standing cat}) in C-GQA. This highlights the strict dependency of the feasibility scores estimated by our model to the particular dataset, with the impossibility to capture polysemy if the dataset does not account for it. These limitations can be overcomed by integrating external information from knowledge bases \cite{liu2004conceptnet} and language\cite{wang2019language}.

{\myparagraph{Retrieving compositions in the open world.} }
In the OW-CZSL scenario there is no limitation in the output space of the model. Thus, we can predict arbitrary state-object compositions at test time and, eventually, retrieve the closest images to arbitrary concepts. Here we explore the latter scenario and we check which images are the closest to the embeddings of random compositions for state and objects not present in the original datasets. The results are shown in Figure \ref{fig:retrieval} for MIT-States (top) and C-GQA (bottom). When the composition is feasible (\eg \textit{calm elephant}, \textit{cut orange}) the model retrieves an image depicting the exact concept. When the composition is inexact for the real world (\eg \textit{closed} vs \textit{folded bike}, \textit{scratched} vs \textit{broken tomato}, \textit{creased} vs \textit{wrinkled dog}) the model still retrieves reasonable images, showing its ability to capture the underlying effect that a state is supposed to have on the particular object it refers. Finally, when the composition has an unclear meaning, the model tends to retrieve images containing both state and objects, even if present in isolation. This is the case of \textit{asphalt bench}, where the \textit{bench} is close to the an \textit{asphalt} road, and of \textit{porcelain table}, where the image shows a \textit{table} with \textit{porcelain} crockery. {Clearly, our model is not perfect and may retrieve images less representative of the queried concept. An example is \textit{blue coffee}, where the retrieved image contains a cup with blue lines}.

\input{sections/retrieval}

%% file: sections/cross-dataset.tex
 \subsection{Cross-dataset results}
 \label{sec:cross}
{\setlength{\tabcolsep}{8pt}
\renewcommand{\arraystretch}{1.0}
\begin{table}[t]
    \centering
    \resizebox{\linewidth}{!}{{\begin{tabular}{l| c c c | c c c }
    \textbf{Training}  & \multicolumn{3}{c}{\textbf{MIT-States}}& \multicolumn{3}{c}{\textbf{C-GQA}}\\
    \textbf{Test}& \multicolumn{3}{c}{\textbf{C-GQA}} & \multicolumn{3}{c}{\textbf{MIT-States}}\\
                                                &S   & U    & HM       
                                                &S   & U    & HM \\\hline
    SymNet\cite{symnet}             
    &  6.5 &    0.93  &0.83
     & 0.44& 0.21      &0.10   \\
    CGE$_\text{ff}$   
    & 6.3 & 1.1      & 1.0  
     & 0.38 &    0.21   &    0.13\\
    \ours$^\text{CW}_\text{ff}$ 
    & 6.2 & 1.1 & 1.0
     & 0.91 &    0.33   &0.15 \\ 
   CGE    
   & 6.3 & 1.4      &1.1   
     & \textbf{1.5} &    0.19  &    0.14\\
          \ours$^{\text{CW}}$       
        & \textbf{7.3} & 1.7     & 1.1   
     & 1.0  & 0.31      &  0.20 \\\hline 
    CompCos  
    &6.3   &2.5       &1.5  
    &0.59   &\textbf{0.49}       &0.17\\
    \ours$_\text{ff}$   
    & 5.5  &  \textbf{3.2}     &  1.6 
     & 0.80 &  0.31     &  0.19 \\
    \ours 
    &  \textbf{7.3}     &  {3.0} & \textbf{2.6}   
     &   1.3&  0.30     &   \textbf{0.23} \\
    \end{tabular}}}
    \vspace{1pt}
    \caption{{\textbf{Cross-dataset CZSL results} on MIT-States compositions from C-GQA models and vice-versa. We measure best seen (S) and unseen accuracy (U), and best harmonic mean (HM) between the two.}}
    \myvspace{-10pt}
    \label{tab:cross}
\end{table}
}

In Section \ref{sec:exp-cw} and Section \ref{sec:exp-ow} we tested CZSL models in standard scenarios, where training and test images belong to the same data collection. However, since the final goal is to perform CZSL in the wild (\eg on web images, robotics applications), we would like such models to show robustness to distribution-shifts, \ie changes between training and test distributions. For these reasons, in this section we benchmark CZSL methods in cross-dataset experiments. 

To perform such experiment, we consider two datasets, MIT-States and C-GQA. The two datasets share 149 objects and 68 states but have different acquisition conditions: while C-GQA contains images of objects cropped from natural images (with eventual loss in resolution), MIT-States contains images downloaded from the web (where the target object is not centered and with possible label noise). For the experiments, we train on the standard training set of one dataset and test on the other. For testing, we consider all the images of the second dataset where the label is a composition of states and objects included in the first. This amounts on testing with 156 seen and 579 unseen compositions on MIT-States, and 131 seen and 687 unseen compositions on C-GQA. Since there is no standard validation set, we take the best models in the OW-CZSL setting of Table \ref{tab:sota-open}, directly applying them on the cross-dataset scenario.

We report the results in Table \ref{tab:cross}, measuring them in terms of best seen accuracy (S) best unseen accuracy (U), and best harmonic mean of the two (HM). 
A first highlight from the table is the much lower accuracy of all models in predicting seen compositions. For instance, CGE performs 1.5\% on MIT-States and 6.3\% on C-GQA vs 32.4\% and 32.7\% on the same datasets in OW-CZSL (Table \ref{tab:cross}). This confirms that there is a significant distribution-shift between the two domains. Despite the shift, \ours\ shows the best overall results in both settings, being always the best in HM and either best or second-best in the other metrics. In detail, when training on MIT-States and testing on C-GQA, our model shows good discriminability on seen classes (7.3\%), being on par with its closed-world counterpart and superior to other closed-world models, such as SymNet (6.5\%) and CGE (6.5\%). On unseen classes, all OW-CZSL methods shows better results, with \ours\ (3.0\%) and \ours$_\text{ff}$ (3.2\%) achieving the best results, almost doubling the performance of the best closed-world model (CGE, 1.7\%). \ours\ shows the best results overall, with 2.6\% HM vs 1.5\% of the best OW-CZSL competitor (CompCos) and 1.1\% of the best closed-world one (CGE and \oursCwEe).

Results are consistent when training on C-GQA and testing on MIT-States, with our model achieving competitive seen accuracy (1.3\% vs 1.5\% of CGE), and unseen accuracy (0.30\% vs 0.31\% of \oursCwEe) with closed-world models. With respect to open-world models, CompCos shows remarkable unseen accuracy (0.49\%) but much lower discriminability on seen compositions (0.59\%). Overall, our model still achieves the best tradeoff between discriminating seen and unseen compositions, achieving 0.23\% HM vs 0.20\% of the best closed-world model (\oursCwEe) and 0.17\% of CompCos. These results show that the efficacy of our model is not restricted to standard OW-CZSL but generalize across distributions. Nevertheless, the overall results of all models are much lower than the ones without distribution-shift, and future works might specifically focus on improving CZSL across domains.

%% file: sections/retrieval.tex
\vspace{-10pt}
{\subsection{Image-retrieval from compositions: results and limitations} 
\label{sec:retrieval}}
{\setlength{\tabcolsep}{3pt}
\renewcommand{\arraystretch}{1.0}
\begin{table}[t]
    \centering
    \resizebox{\linewidth}{!}{{\begin{tabular}{l| c c  | c c |  c c  | c c | c c c }
    \textbf{Training}& \multicolumn{4}{c}{\textbf{MIT-States}}& \multicolumn{4}{c|}{\textbf{C-GQA}}&\multicolumn{3}{c}{\multirow{2}{*}{\textbf{Average}}}\\
        \textbf{Test}& \multicolumn{2}{c}{\textbf{MIT-Sta.}}& \multicolumn{2}{c|}{\textbf{C-GQA}}& \multicolumn{2}{c}{\textbf{MIT-Sta.}}& \multicolumn{2}{c|}{\textbf{C-GQA}}&&&\\
                               
                                                &S   & U &S   & U &S   & U &S   & U & S &U &HM\\\hline
  
    CGE$_\text{ff}$     
     &23.3 &22.5
     &9.9 &4.2
     &14.1 &6.7
     & 12.2&3.8
     &14.9&9.3&10.9\\
    \ours$^\text{CW}_\text{ff}$
     &26.8 &\textbf{30.8}
     &13.7 &4.8
     &11.5 &5.4
     &11.1 &2.9
     &15.8&\textbf{11.0}&11.9\\ 
   CGE   
     &25.5 &24.8
     &12.2 &\textbf{5.1}
     &\textbf{16.0} &5.4
     &12.5 &\textbf{4.6}
     &16.6&9.9&11.7\\
          \ours$^{\text{CW}}$      
     &\textbf{30.5} &26.0
     &\textbf{15.3} &\textbf{5.1}
     & 9.6&\textbf{7.3}
     &\textbf{12.8} &3.6
     &\textbf{17.1}&10.5&\textbf{12.4}\\\hline
    CompCos   
     &26.2 &25.0
     & \textbf{13.7}&3.9
     &\textbf{15.4} &6.0
     &\textbf{12.6} &\textbf{3.6}
     &\textbf{17.0}&9.6&11.5\\
    \ours$_\text{ff}$
     &\textbf{28.0} &\textbf{30.0}
     & 12.2&4.1
     & 9.0&\textbf{6.9}
     &10.5 &2.8
     &14.9&\textbf{11.0}&11.8\\
    \ours 
     &25.0 &27.3
     & \textbf{13.7}&\textbf{6.1}
     &12.8 &5.7
     &12.4 &3.3
     &16.0&10.6&\textbf{11.9}\\
    \end{tabular}}}
    \vspace{1pt}
    \caption{{\textbf{Compositional retrieval results} on MIT-States and C-GQA, including cross-dataset tests. We report the top-1 accuracy on seen (S) and unseen (U) compositions.}}
    \myvspace{-10pt}
    \label{tab:retrieval}
\end{table}
}

In the previous section, we reported a qualitative study of retrieving compositions of arbitrary objects and states using our OW-CZSL model. A natural question is whether the learned compositional embeddings can be used to reliably retrieve images of the corresponding concepts. We conducted this study on MIT-States and C-GQA performing the experiments both within and across dataset, following the setup of Section \ref{sec:cross} for the latter. In this setting, we benchmark the best methods of Table \ref{tab:sota-open} that explicitly learn compositional embeddings: CGE, \oursCwEe\ and their frozen counterparts among the closed-world CZSL methods, and all the OW-CZSL ones (\ie CompCos, \ours$_\text{ff}$ and \ours).  We measure the results according to accuracy on seen compositions (S), unseen compositions (U) and their average across all settings, together with their average harmonic mean (HM). Since focusing only on seen compositions may lead to more discriminative compositional embeddings, we highlight the results of the best models trained in the closed- and open-world settings separately.

We report the results of our experiment in Table \ref{tab:retrieval}. As a first observation, there is no clear winner across all settings. For instance, \oursCwEe\ achieves the best results on seen compositions across all settings but the cross-dataset experiment C-GQA to MIT-States, where it achieves poor performance (9.6\% vs 16.0\% of CGE). Similarly, \ours\ shows the best unseen accuracy on the cross-dataset experiment MIT-States to CGQA, but lower than its non-finetuned counterpart when testing on MIT-States (\ie 30.0\% vs 27.3\% for within, and 6.9\% vs 5.7\% for cross dataset experiments). CompCos shows good retrieval results when trained on C-GQA, achieving the best accuracy on seen classes among OW-CZSL models and either best or second-best on unseen ones. Overall, our \oursCwEe\ and \ours\ shows the best trade-off between seen and unseen accuracy in all setting, achieving a slightly better average HM than the best competitors (\ie 11.9\% \ours\ vs 11.8 \ours$_\text{ff}$ and 11.5\% CompCos, 12.4\% of \oursCwEe\ vs 11.9\% of \oursCw\ and 12.4\%), with cosine-based classifiers always surpassing their non-cosine counterpart in all metrics (\ie CGE vs \oursCwEe\ and CGE$_\text{ff}$ vs \oursCw). 

These results show that better CZSL results do not necessarily lead to better compositional embeddings. For instance, while in Table \ref{tab:sota-open} \ours\ and clearly surpasses CompCos (\ie 0.78 AUC on C-GQA, 2.3 on MIT-States vs 0.39 and 1.6 respectively) the margins in the retrieval setting are smaller (\eg 10.6\% vs 9.6\% on average unseen accuracy) and the best model may even change (\eg 17.0\% average seen accuracy for CompCos vs 16.0\% of \ours). Refining the  compositional embeddings by exploiting advances on metric learning (\eg \cite{cakir2019deep,roth2020revisiting}) may improve results on both compositional-retrieval and standard CZSL.

%% file: sections/conclusion.tex
In this work, we address the compositional zero-shot learning (CZSL) problem, where the goal is recognizing unseen compositions of seen state-object primitives. We focus on the  
open world compositional zero-shot learning (OW-CZSL) task, where all the combinations of states and objects could potentially exist. 
We propose a way to model the feasibility of a state-object composition by using the visual information available in the training set.  This feasibility is independent of an external knowledge base and can be directly incorporated in the optimization process.
We propose a novel model,{ \ours}, that models the relationship between primitives and compositions through a graph convolutional neural network. \ours\ incorporates the feasibility scores in two ways: as margins for a cross-entropy loss and as weights for the graph connections. Experiments show that our approach is either superior or comparable to the state of the art in closed world CZSL while improving it by margin in the open world setting. In the future, we plan to study new ways of computing our feasibility scores, as well as the effectiveness of our compositional graph in other problems with interacting sets (\eg human-object interaction, action recognition). {We also plan to investigate different GCN architectures (\eg \cite{gcc}) to improve our compositional graph embeddings.}  Finally, it would be interesting to merge the advances in ZSL for recognizing unseen semantic concepts (\ie new primitives) with CZSL ones in modeling the interactions among primitives, to develop models that can extrapolate compositional capabilities to not only unseen compositions of known concepts, but also to compositions of unseen objects and states.